\documentclass[final,onecolumn,1p]{elsarticle} 
\usepackage{graphicx}
\usepackage{amsthm}
\usepackage{amssymb} 
\usepackage{amsmath} 
\usepackage{threeparttable}
\usepackage{txfonts} 
\usepackage{setspace}
\usepackage{bm}
\usepackage{enumitem}
\usepackage{caption}
\usepackage{float}

\usepackage{multirow} 
\usepackage{diagbox} 
\usepackage{threeparttable} 

\usepackage{algorithm}
\usepackage{algorithmic}

\usepackage{wrapfig}
\usepackage{color}

\usepackage{lineno,hyperref}
\modulolinenumbers[5]

\begin{document}

\begin{frontmatter}

\title{Improving generative adversarial network inversion via fine-tuning GAN encoders}

\author[a,b]{Cheng Yu}
\ead{disanda@cqut.edu.cn}

\author[b]{Wenmin Wang\corref{cor1}}
\ead{wmwang@must.edu.mo}

\author[c,d]{Roberto Bugiolacchi}
\ead{rbugiolacchi@must.edu.mo}

\address[a]{School of Artificial Intelligence, Chongqing University of Technology, Chongqing , China}

\address[b]{School of Computer Science and Engineering, Macau University of Science and Technology, Macau, China}

\address[c]{\noindent State Key Laboratory of Lunar and Planetary Sciences, Macau University of Science and Technology, Macau, China}

\address[d]{Earth Sciences, University College London, London WC1E 6BT, United Kingdom}

\cortext[cor1]{Corresponding author.}

\begin{abstract}
Generative adversarial networks (GANs) can synthesize high-quality (HQ) images, and GAN inversion is a technique that discovers how to invert given images back to latent space. While existing methods perform on StyleGAN inversion, they have limited performance and are not generalized to different GANs. To address these issues, we proposed a self-supervised method to pre-train and fine-tune GAN encoders. First, we designed an adaptive block to fit different encoder architectures for inverting diverse GANs. Then we pre-train GAN encoders using synthesized images and emphasize local regions through cropping images. Finally, we fine-tune the pre-trained GAN encoder for inverting real images.  Compared with state-of-the-art methods, our method achieved better results that reconstructed high-quality images on mainstream GANs. Our code and pre-trained models are available at: https://github.com/disanda/Deep-GAN-Encoders.
\end{abstract}


\end{frontmatter}

\section{Introduction}

With the burgeoning advance of deep neural networks, generative adversarial networks (GANs) \cite{GAN,DCGAN,PGAN} can synthesize high-quality (HQ) images. For instance, BigGAN \cite{Big-GAN} can synthesize HQ images with sizes of $128\times128$, $256\times256$, and $512\times512$. It performs label-supervised training on ImageNet \cite{ImageNet}, which is a large-scale dataset with 1000 labels. Moreover, StyleGAN and upgrade versions \cite{Style-GAN,StyleGAN2} can generate high-quality images, such as cars, cats, and horses with resolutions of $256\times256$ on LSUN \cite{lsun}, and high-quality faces with resolutions of up to $1024\times1024$ pixels on CelebA \cite{celeba}.

GANs are well-known for their impressive performance in image synthesis, suggesting that there is potential for developing GAN-based technologies for real image editing \cite{InterfaceGAN,RFM}. Despite the success of state-of-the-art GANs like BigGAN and StyleGANs, which use layer-wise latent representation to accurately synthesize images.

\begin{figure}[htbp]
\centering
\includegraphics[width=\linewidth]{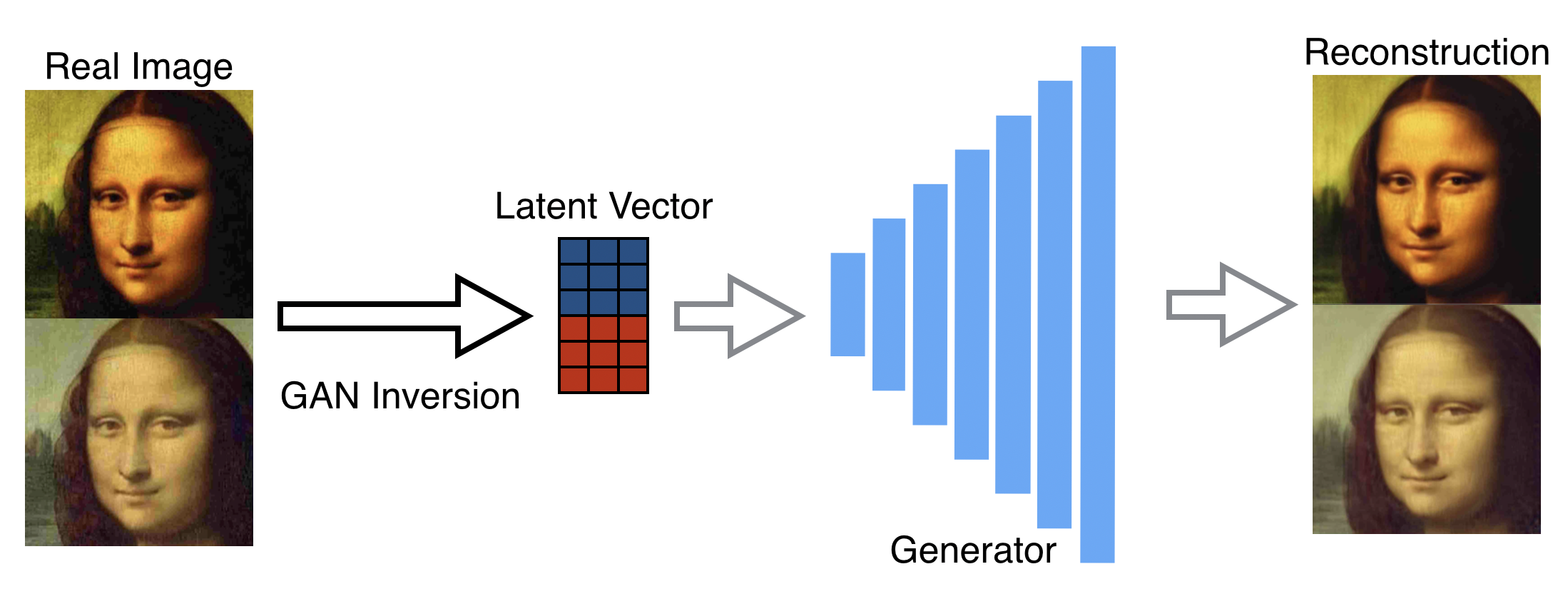} 
\caption{Overview of GAN inversion, Inverting real images to latent vectors is called GAN inversion (indicated by black arrow). Using the latent vectors as input to the generator from pre-trained GAN, we can generate reconstructions (indicated by black arrows).}
\label{gan-inversion}
\end{figure}

As shown in Fig. \ref{gan-inversion}, inverting given images into latent space is a technique known as GAN inversion \cite{GAN-inversion}. GAN-based applications utilize inverted latent vectors to represent edited images or videos \cite{RiDDLE}. Previous methods \cite{InvertGAN,FastEncoder,Perarnau2016} have attempted to design an encoder for GAN inversion. Such methods perform well for shallow networks and low-quality images (below $256\times256$), but they are not effective for GANs that fail to invert HQ images into the latent space.

From the perspective of optimizing parameters, there are three primary approaches for GAN inversion. : (1) \textbf{Optimizing latent vector}, which directly optimizes latent vectors \cite{img2stylegan}; (2) \textbf{Training encoder}, which involves training an encoder to generate latent vectors \cite{E2Style,ALAE}; (3) \textbf{Hybrid method}, which combines the previous two methods by first training an encoder and then optimizing the latent vectors \cite{InDomainG}. 

From the perspective of optimizing objectives, while \cite{RiDDLE,E2Style,pSp} add facial identity features to improve face inversion performance, they can not be generalized to other non-face images. Consequently, these methods have limited performance, and generalizability to other GANs since they are only designed for StyleGANs.

Unlike existing methods, we utilize cropped images instead of facial identities to enhance method generalization. Meanwhile, we are the first to boost GAN inversion by fine-tuning the pre-trained GAN encoders. The detailed comparisons are shown in Table \ref{opt}:

\begin{table*}[htbp]
\small
\centering
\caption{Overview of Optimization Parameters and Objectives for Different GAN Inversion Methods.}
\label{opt}
\renewcommand\arraystretch{1.2} 
\setlength{\tabcolsep}{1.9pt} 
\newcommand{\tabincell}[2]{\begin{tabular}{@{}#1@{}}#2\end{tabular}} 
\begin{tabular}{|c|ccc|cc|}
\hline \hline
\multirow{2}{*}{Method} & \multicolumn{3}{c|}{Parameters}                                   & \multicolumn{2}{c|}{Objectives}     \\ \cline{2-6} 
 & \multicolumn{1}{c|}{Latent Vector} & \multicolumn{1}{c|}{Encoder(Training)} & Encoder(Fine-tuning) & \multicolumn{1}{c|}{Identity Feature} & Cropping Image \\ \hline
ALAE\cite{ALAE}                   & \multicolumn{1}{c|}{}          & \multicolumn{1}{c|}{$\bullet$} &           & \multicolumn{1}{c|}{}          &           \\ \hline
In-domain\cite{InDomainG}               & \multicolumn{1}{c|}{$\bullet$} & \multicolumn{1}{c|}{$\bullet$} &           & \multicolumn{1}{c|}{}          &           \\ \hline
pSp\cite{pSp}             & \multicolumn{1}{c|}{}          & \multicolumn{1}{c|}{$\bullet$} &           & \multicolumn{1}{c|}{$\bullet$} &           \\ \hline
E2Style\cite{E2Style}                & \multicolumn{1}{c|}{} & \multicolumn{1}{c|}{$\bullet$} &           & \multicolumn{1}{c|}{$\bullet$} &           \\ \hline
RiDDLE\cite{RiDDLE}                  & \multicolumn{1}{c|}{} & \multicolumn{1}{c|}{$\bullet$} &           & \multicolumn{1}{c|}{$\bullet$} &           \\ \hline
Ours                    & \multicolumn{1}{c|}{$\bullet$} & \multicolumn{1}{c|}{}          & $\bullet$ & \multicolumn{1}{c|}{}          & $\bullet$ \\ \hline \hline
\end{tabular}
\end{table*}

\begin{figure*}[htbp]
\centering
\includegraphics[width=\linewidth]{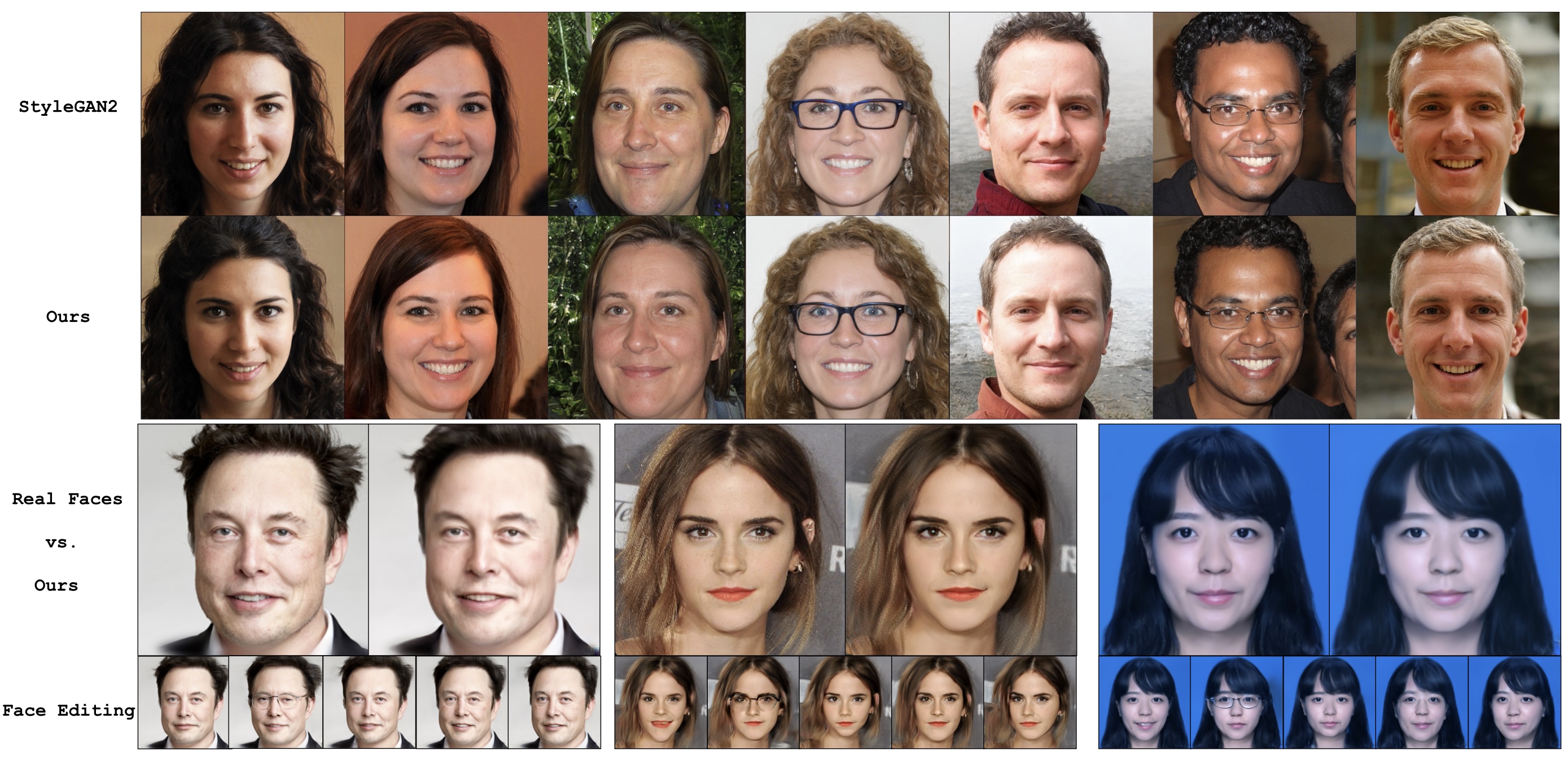} 
\caption{The 1st row displays face images synthesized by StyleGAN2 (FFHQ 1014×1024) \cite{StyleGAN2}. In the 2nd row, we present our reconstructions. Using our method, the 3rd row shows three real faces (on the left) and their reconstructions (on the right). Our method accurately reproduces the original faces. The 4th row demonstrates the ability of our method to edit faces using five interpretable latent directions using \cite{RFM}: mouth, eyeglasses, younger, older, and pose.}
\label{fig1}
\end{figure*}

To invert various GANs, we design an adaptive encoder block that can be regarded as a universal GAN encoder architecture. Our method not only reconstructs real faces with high fidelity but also maintains their identity features and semantic attributes, particularly in the case of human faces (as shown in Fig. \ref{fig1}).

We transform pre-trained GAN into an auto-encoder, an encoder-decoder architecture. Next, we regard pre-trained generators as decoders and design corresponding encoders in GANs. Our contributions are summarized as follows:

\begin{itemize}

\item We designed an adaptive encoder block for GAN inversion. The block helps us to build different GAN encoders. Our method is the first effort to create diverse GAN encoders to match mainstream GANs via the adaptive block.

\item 
To perform GAN inversion on high-quality (HQ) images, we improved the loss function by cropping images to obtain image attentions that highlight the key areas of these images. This attention can be applied to both center-aligned and misaligned images. Furthermore, we further enhanced the loss function by incorporating the structural similarity metric.

\item Unlike current methods \cite{RiDDLE,pSp,InDomainG}, which rely on hybrid approaches for inverting real images (i.e., training an encoder and then optimizing the latent vector), we fine-tune the pre-trained encoder to achieve superior inversion performance, e.g., real face reconstructions. Besides, our method enables real face editing via label-based latent directions, which outperforms state-of-the-art performance compared to the current method \cite{InterfaceGAN}.

\end{itemize}

\section{Related Work}
\subsection{GAN Inversion}

GAN inversion \cite{GAN-inversion} aims to invert given images to latent vectors and then faithfully reconstruct images using well-trained GANs. Meanwhile, GAN inversion methods could coordinate usable latent directions to edit given images. Various GAN-based applications depend on GAN inversion, such as face editing \cite{e4e,GANspace} and text-to-image \cite{StyleClip}.

GANs can be regarded as a distinct form of auto-encoder, with the generator serving as the decoder and the discriminator as the encoder \cite{FastEncoder}. Unlike conventional encoders, the discriminator in GANs solely distinguishes between real and fake images and cannot encode images into latent codes.

This above limitation prevents GANs from encoding output images into latent space. The adversarial auto-encoder (AAE) \cite{aae} and its upgraded version \cite{ALAE} integrate a variational auto-encoder (VAE) [23] with GAN to realize a GAN encoder. However, AAE exhibits limited performance in GAN inversion. This is because AAE is designed to learn presentations on training dataset images, not for given images.

Here, given images are synthesized or real images. AAE focuses on mapping the training data distribution into a latent distribution for synthesized images, rather than inverting real images. Most real images are not part of the training datasets, and cannot be accurately reconstructed by AAE. Therefore, GAN inversion is an essential part of the learning representation in GANs.

Latent vectors, also referred to as latent codes or latent variables, are sampled from the latent space. Traditional approaches \cite{Perarnau2016,Bi_GAN} rely on shallow encoders to invert low-quality images into latent space. However, inverting high-quality images into latent space poses a challenge due to their large size, which can be up to $1024\times1024$ pixels.

If the dimensionality of latent vectors is too small (e.g., $\leq 256$), mode collapse or spatial entanglement can occur in the latent representation. The size gap between images and latent vectors makes GAN inversion even more difficult. Although previous methods \cite{sampleGAN,Image-Manifold} reduced spatial entanglement by interpolating latent space on manifold directions, they only work in a few cases with shallow GANs.

In StyleGANs, the layer-wise style latent vector ($\mathbf{w}$) is in each convolution layer besides the typical latent vector which is the first layer input ($\mathbf{z}$). Based on the layer-wise manner, Image2StyleGAN \cite{img2stylegan} directly optimizes $\mathbf{w}$ for GAN inversion (Latent code method). However, the method requires a significant number of training operations (around 3,500 iterations per image) for most given images, making it less efficient when inverting real images. Therefore, we use an encoder for the GAN inversion.

For the StyleGAN inversion, In-Domain GAN \cite{InDomainG} trained an encoder to invert dataset images into $\mathbf{w}$, and In-Domain GAN inverted a given image via optimizing $\mathbf{w}$ based on the pre-trained encoder. ALAE \cite{ALAE} use an auto-encoder framework for StxyleGAN inversion. To improve StyleGAN inversion, pSp \cite{pSp} added face identity loss, and E2Style \cite{E2Style} upgrades pSp by simplifying the style mapping layer. RiDDLE \cite{RiDDLE} enhances face identity features to edit real face images.

For BigGAN inversion, drawing on the concept of BiGAN \cite{Bi_GAN}, BigBiGAN \cite{BigBiGAN} improves the discriminator to judge image-latent pairs (rather than just judging images as in standard GANs) and trains the encoder with an upgraded BigGAN to obtain the BigGAN image reconstructions.

While existing methods have inverted real images to latent vectors, these methods are specific to StyleGANs for real aligned faces. We improve the GAN inversion performance in terms of generalization.

In addition, interpretable directions can be discovered on latent space using supervised methods \cite{InterfaceGAN,RFM},  which also can be discovered via unsupervised methods \cite{GANspace,LatentCLR}. Interpretable directions are widely utilized for image editing and video generation.

\subsection{Similarity Evaluation}

Mean square error (MSE) and cosine similarity (COS) are used to measure the similarity of images and latent vectors in previous training loss functions \cite{pSp,InDomainG,ALAE}. However, the two methods below are better suited for measuring image similarity:

\vspace{5pt}
\textbf{LPIPS}. To measure the perceptual similarity of images, learned perceptual image patch similarity (LPIPS) requires a pre-trained model \cite{perceptual} (e.g., VGG \cite{VGG} or AlexNet \cite{AlexNet}), which is trained on ImageNet \cite{ImageNet} for classification. LPIPS extracts multi-layer features to evaluate image similarity between images and their reconstructions. To evaluate the similarity of HQ images more consistently, we also apply SSIM.
 
\vspace{5pt}
\textbf{SSIM}. Reconstructed images are subject to inevitable distortions in deep neural networks, such as passing the down- or up-sampling layer, including the convolutional, normalization, pooling, and activation layers. Any of these steps have the potential of degrading the reconstruction quality. We consider adding structural similarity (SSIM) \cite{SSIM} to measure jointly the perceptual feature similarity. SSIM has high computational efficiency and does not require pre-training models.

\vspace{5pt}
Current methods for evaluating image similarity are limited, especially when dealing with HQ images. To improve the performance of GAN inversion, it is essential to incorporate a more diverse set of similarity metrics. Despite LPIPS being widely used, SSIM offers higher computational efficiency and can be an auxiliary loss function term for training and fine-tuning encoders. Therefore, we propose a loss function that combines multiple similarity metrics for GAN inversion.

\vspace{5pt}

\textbf{Grad-CAM}. For HQ images, the evaluation of image similarity is not sufficient. While \cite{RiDDLE,E2Style,pSp} attempt to augment similarity assessment by integrating facial identity information, they are confined to face images. To address more types of images, we crop images to highlight objects and evaluate image similarity using the cropped images to enhance inversion performance. However, some misaligned image objects are in random positions, and these images are hard to crop. Consequently, we need a method to crop misaligned images.

\vspace{5pt}
The class activation mapping (CAM) \cite{CAM} is to locate object positions within images using a classification model pre-trained on ImageNet \cite{ImageNet}. However, to obtain object locations, CAM needs to train the CAM model again. Gradient-weighted class activation mapping (Grad-CAM) upgrades CAM by utilizing a pre-trained model gradient at the pixel level and does not require additional training. We utilize Grad-CAM to crop misaligned images to focus more on the image objects.

\section{Method}

\textbf{Overview}. In this section, we propose an adaptive block that can be used for different architectures of GAN encoders. We first introduce our encoder-decoder symmetric architecture, and the adaptive encoder block, followed by the cropping attentions, and the loss function used for training and fine-tuning.

\textbf{Symbol denotation}. For the model part, we denote the GAN generator as $G$. The GAN encoder is denoted as $E$. For conditional GANs (e.g., BigGAN), the label vector is denoted by $\mathbf{c}$.

For the image part, we denote the real image as $\mathbf{y}$, and its reconstruction is denoted as $\mathbf{y'}$. The GAN-synthesized image is denoted by $\mathbf{x}$, and its reconstruction is denoted as $\mathbf{x'}$. $AT1$ and $AT2$ denote image cropping functions, while $\mathbf{x_1}$ and $\mathbf{x_2}$ refer to cropped synthesized images that emphasize image objects. The cropped reconstructions are represented by $\mathbf{x_1'}$ and $\mathbf{x_2'}$.

For the latent space of most shallow GANs, the latent vectors are not layer-wise and go directly from the first layer. So we denote the latent vector as $\mathbf{z}$ of the given image ($\mathbf{x}$ or $\mathbf{y}$). $\mathbf{z'}$ denotes the inverted latent vector of the reconstructed image ($\mathbf{x'}$ or $\mathbf{y'}$).

For StyleGAN latent space, its first layer input is a constant and we denote it as $\mathbf{z_c}$ instead of $\mathbf{z}$. There are two additional layer-wise vectors in StyleGAN: the style vector ($\mathbf{w}$) and the noise vector ($\mathbf{z_n}$). So the latent vectors of StyleGAN are ($\mathbf{w}$, $\mathbf{z}_c$, $\mathbf{z}_n$) and their inverted vectors as: ($\mathbf{w}', \mathbf{z}_c', \mathbf{z}_n'$).

Here, the GAN inversion task aims to obtain inverted latent vectors (e.g., $\mathbf{w'}$ of StyleGAN) from the GAN encoder. Let $\mathbf{w'}$ get the reconstruction abilities of $\mathbf{w}$ while ensuring that $\mathbf{x'}$ (or $\mathbf{y'}$) is as similar as possible to $\mathbf{x}$ (or $\mathbf{y}$).

We demonstrate the data flow of our method in Fig. \ref{fig_flow}:

\begin{figure}[htbp]
\centering
\includegraphics[width=0.95\linewidth]{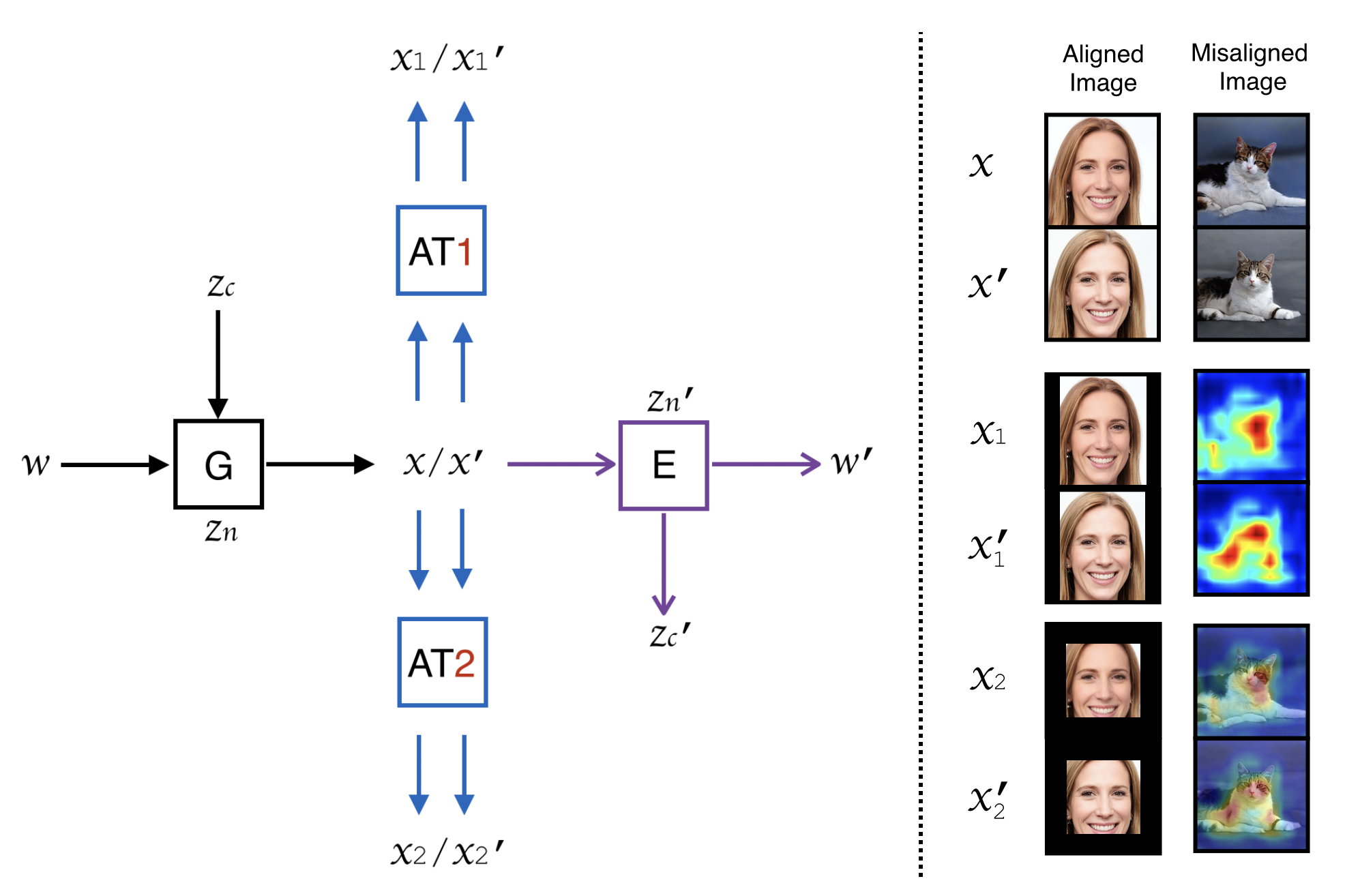} 
\caption{Inversion data flow for StyleGANs consists of 3 steps. In \textbf{Step 1}, we input latent vectors ($\mathbf{w}$, $\mathbf{z_c}$) with their inverted vectors ($\mathbf{w'}$, $\mathbf{z'_c}$) to $G$ to synthesize and reconstruct images ($\mathbf{x}$, $\mathbf{x'}$). In \textbf{Step 2}, we crop $\mathbf{x}$ and $\mathbf{x'}$ to obtain two attention areas that highlight main objects in images. At \textbf{Step 3}, we train $E$ to invert images back to latent vectors.}
\label{fig_flow}
\end{figure}

\subsection{Encoder Architecture} 

We make the architecture of $E$ symmetrical to $G$ by designing an adaptive block. In a symmetric architecture \cite{rethink_encoder_decoder}, the encoder and decoder are equally powerful and mirror each other's design. This encoder architecture guarantees that the features extracted by the encoder work well with the decoder, improving the representation via encoder latent space.

Meanwhile, the symmetric architecture can help preserve the features from the image space to the latent space and back to the image space. Training can be more efficient, which leads to easier convergence.

\begin{figure}[htbp]
\centering
\includegraphics[width=0.55\linewidth]{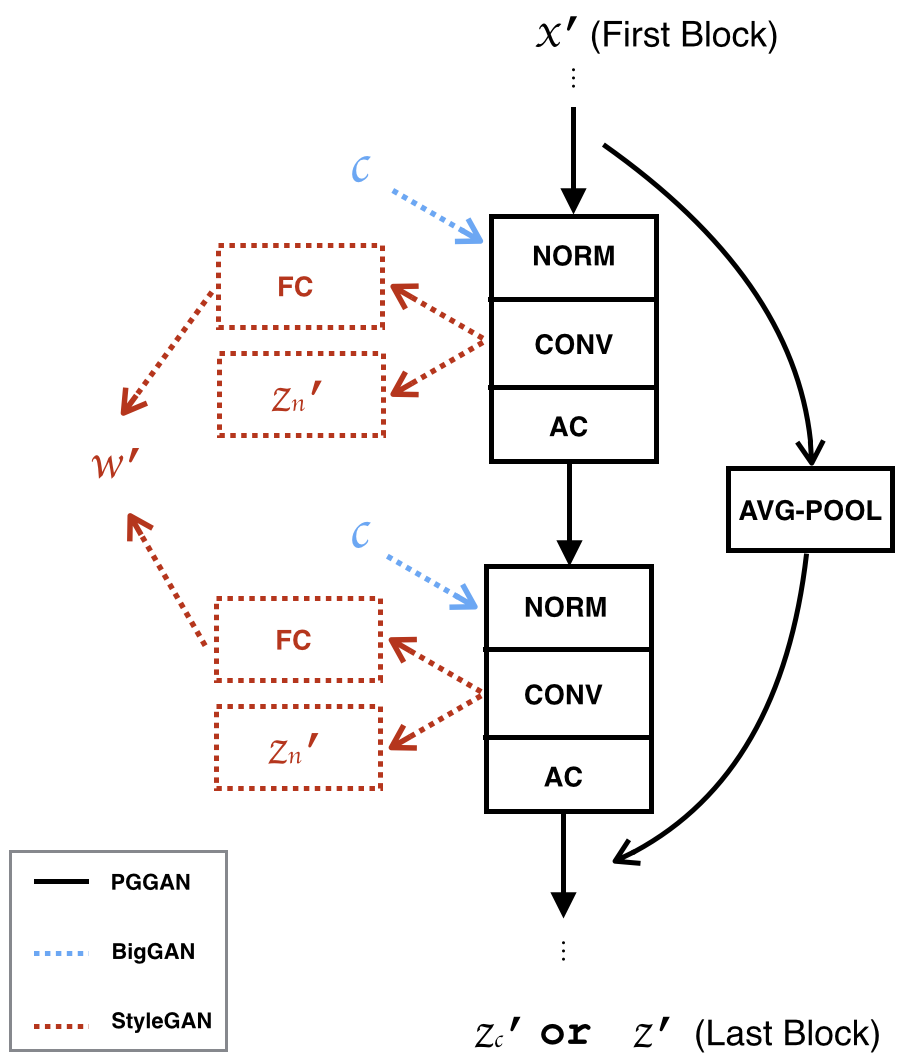} 
\caption{Overview of the adaptive block architecture for GAN encoder. NORM denotes the normalization layer, and AC denotes the activation layer. FC and learnable $\mathbf{z}_n'$ are removed in other PGGAN and BigGAN. We have also added layer-wise label vectors $c$ for BigGAN inversion.}
\label{fig_block}
\end{figure}

The adaptive block is illustrated in Fig. \ref{fig_block}. Each block consists of two convolutional layers (CONV), and each CONV followed by a learnable noise vector ($\mathbf{z}_n'$) and a layer-wise style vector $\mathbf{w}'$. $\mathbf{w}'$ outputs from fully connected layers (FC). Both FC and CONV use an equalized learning rate \cite{PGAN}. 

To perform the symmetric architecture, we designed the encoder blocks to mimic the generator blocks. We adopted the residual bypass as in \cite{ResNet} and incorporated it into each encoder block. 
 
As we add more layers to GAN encoders, training can become slower and accuracy might not improve, or even decline. Our residual blocks address this problem and handle the vanishing gradient issue.

Below are three cases demonstrating how we adapted the encoder block to match different GAN generators:

\vspace{5pt}
\textbf{StyleGANs}. We input latent vectors $(\mathbf{w}, \mathbf{z}_c)$ to $G$ and get $\mathbf{x}$. Next, we invert vectors $(\mathbf{w}', \mathbf{z}_c')$ using $E$. We do not optimize latent vectors here. Instead, we train and fine-tune GAN encoders to get the ideal inverted latent vectors.

\vspace{5pt}
\textbf{PGGAN}. We remove the style- and noise-related vectors (i.e., $\mathbf{w}'$ and $\mathbf{z}_n'$), and remove style FC in each block. At the last block, we changed the last CONV to an FC to adapt to PGGAN. Distinct from StyleGANs, the latent vector of PGGAN is $\mathbf{z} \in \mathbb{R}^{512}$ from the first layer.

\vspace{5pt}
\textbf{BigGAN}. We also remove $\mathbf{w}'$, $\mathbf{z}_n'$ with the relevant layers from each block. We replace the instance norm \cite{IN} with the conditional batch normalization (CBN). $\mathbf{c} \in \mathbb{R}^{256}$ denotes ImageNet labels that are input into $G$ to synthesize the label-based image. Similar to $\mathbf{w}$, $\mathbf{c}$ in BigGAN is a layer-wise vector and its imitated vector is $\mathbf{c}'$. In the last block of BigGAN, two FC layers output $\mathbf{c}'$ and $\mathbf{z} \in \mathbb{R}^{128}$ separately. We perform the label vectors as one-hot vectors. 

\vspace{5pt}
In the \textbf{Appendix}, we report the architectural details of the encoders used for StyleGANs, PGGAN, and BigGAN. We can modify the blocks to accommodate different latent representations when we use the adaptive block for other GAN encoders, following our above cases.

\subsection{Pre-training Loss Function}

Training the encoder aims to invert synthesized images to the latent space and reconstruct these images using the encoded latent vectors, i.e., $\mathbf{x} \simeq \mathbf{x'}$ as follows:

\begin{equation}
\mathcal{L}_E = \mathcal{L}_{IMG}(\mathbf{x}, \mathbf{x}').
\label{eq1}
\end{equation}

Here, the objective of  $\mathcal{L}_{IMG}(\mathbf{x}, \mathbf{x}')$ is to make the reconstructed image ($\mathbf{x}'$) sufficiently similar to the original image ($\mathbf{x}$).  To train the encoders, we adopt a self-supervised approach based on pre-trained generators. We synthesize training samples by feeding random latent vectors ($\mathbf{z}$) into the generator, $\mathbf{x} \simeq G(\mathbf{z})$. We evaluate the similarity of image $\mathbf{x}$ and its reconstruction $\mathbf{x}' \simeq G(E(G(\mathbf{z})))$ attached with cropped attentions. To ensure comparability, we resize cropped attentions $(\mathbf{x}_1, \mathbf{x}_1')$ and $(\mathbf{x}_2, \mathbf{x}'_2)$  up to the same size of images $(\mathbf{x}, \mathbf{x}')$.

\vspace{5pt}
\textbf{Cropping Attention}. Inverting HQ images to latent space while preserving high-fidelity reconstructed images is a challenging task because the encoding process often leads to the loss of feature information. A common issue is the loss of local features. To address this, we propose to crop the images to emphasize the objects of interest. These cropped images are supplemental samples of the training dataset in each training iteration.

Cropping images directs the encoder's attention to specific areas, which is beneficial when key features are concentrated in particular regions. By emphasizing these areas, the model learns more pertinent features while disregarding irrelevant ones, leading to better performance \cite{aoa}.

We crop two attentions, denoted as AT1 and AT2. The cropped attentions and their reconstructions are: ($\mathbf{x}_1, \mathbf{x}_1'$) and ($\mathbf{x}_2, \mathbf{x}_2'$).  These attentions emphasize image objects. AT1 represents the first cropping, performed as $ \mathbf{x}_1, \mathbf{x}'_1 = AT1(\mathbf{x}, \mathbf{x}')$. Based on AT1, AT2 represents the subsequent cropping: $ \mathbf{x}_2, \mathbf{x}'_2 = AT2(\mathbf{x}, \mathbf{x}')$. Both attentions outstanding key objects in HQ images. Finally, the image loss function is given by:

\begin{equation}
\small
\mathcal{L}_{IMG}(\mathbf{x}, \mathbf{x}') 
\rightarrow 
\mathcal{L}(\mathbf{x}, \mathbf{x}')+ \mu_1 \mathcal{L}(AT_{1}(\mathbf{x}, \mathbf{x}')) + \mu_2 \mathcal{L}(AT_{2}(\mathbf{x}, \mathbf{x}')).
\label{eq3}
\end{equation}

Where $\mu_1$ and $\mu_2$ are hyper parameters. For center-aligned images, we set empirically to:  $\mu_1$ to 0.375 and $\mu_2$ to 0.625.

\begin{figure}[htbp]
\centering
\includegraphics[width=0.55\linewidth]{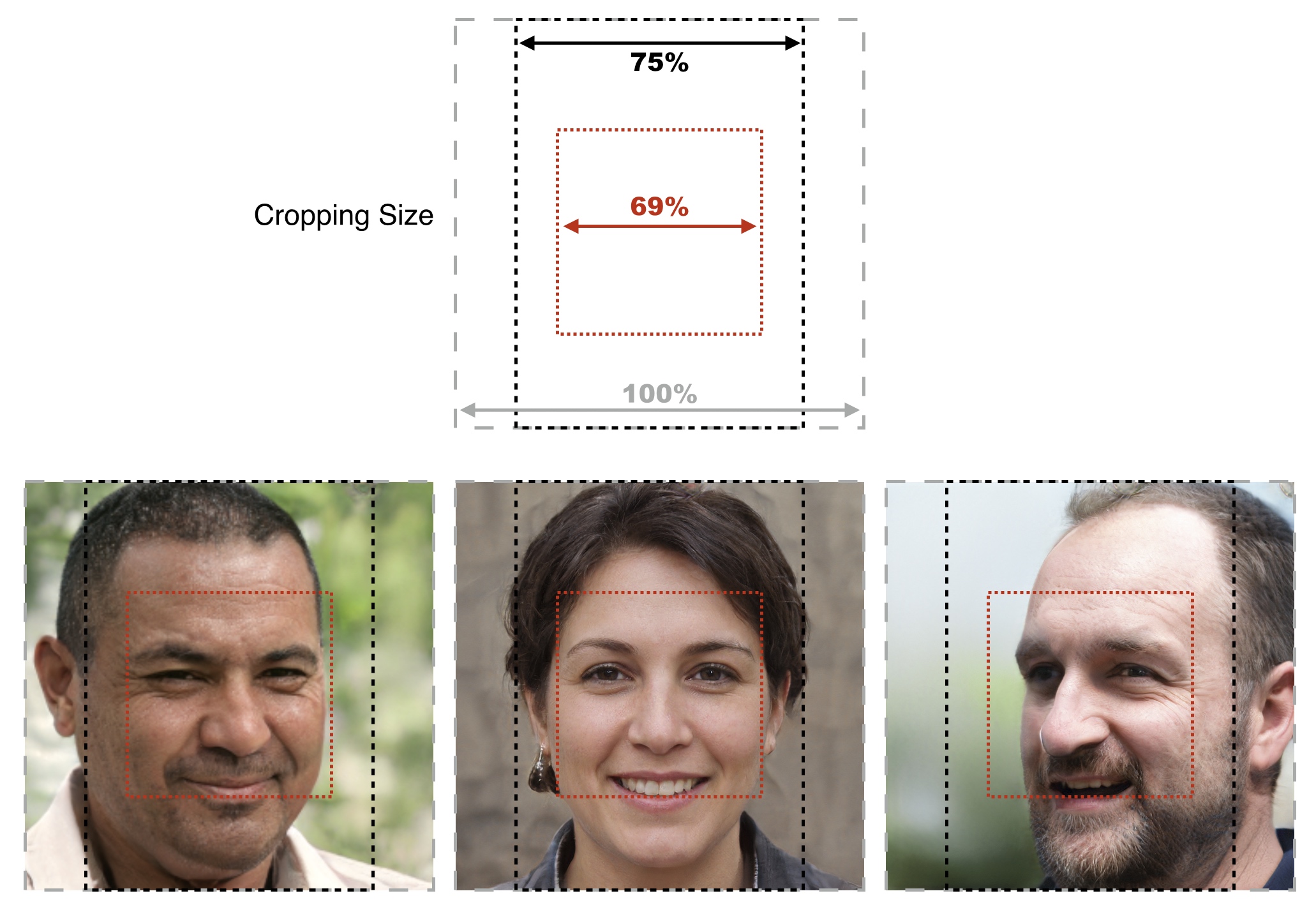} 
\caption{Cropping attention to center-aligned faces. The gray dashed line indicates the boundaries of the original image. Empirically, the black dashed line crops the first attention (AT1) around 0.75$\%$ width, and the red dashed line crops the second attention (AT2) around 0.69 $\%$ width and height from the original figures.}
\label{fig_aoa_1}
\end{figure}

\vspace{5pt}
\textbf{Aligned Image}. For preprocessed images, such as faces from CelebA-HQ \cite{CelebAMask-HQ}, the dataset images are center-aligned, making it easy to crop margin areas to highlight the face objects. As shown in Figure \ref{fig_aoa_1}, we create two attentions for each face: AT1 crops along the black dashed line that contains the whole head, and AT2 crops along the red dashed line that contains the facial features. We determined the cropping margins empirically: AT1 crops $12.5\%$ of the pixels in width (75$\%$), while AT2 crops around $15.5\%$ of the pixels in both width and height ($69\%$).

\begin{figure}[htbp]
\centering
\includegraphics[width=0.8\linewidth]{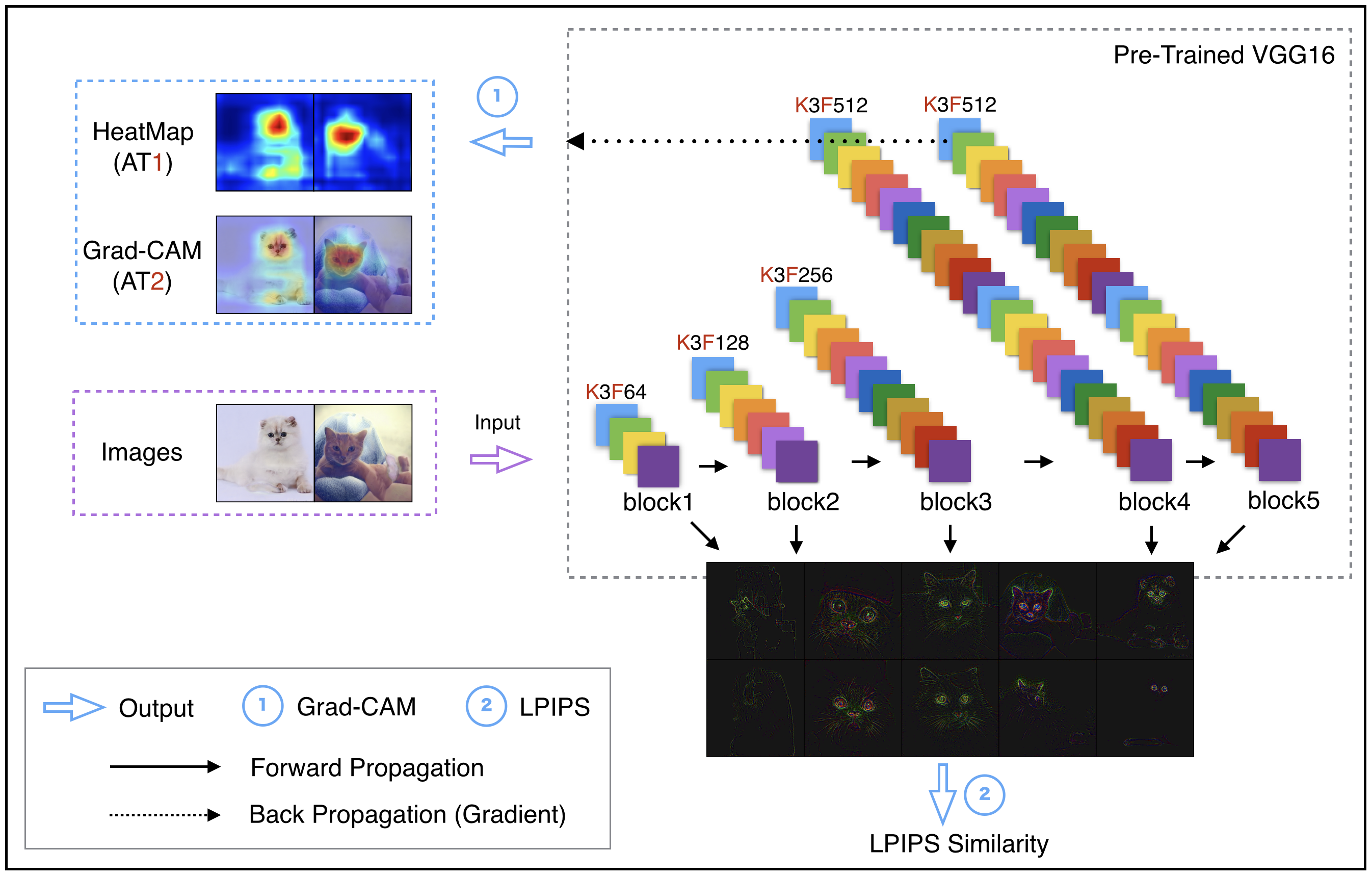} 
\caption{Cropping attention (Grad-CAM-based) for misaligned images. The first attention (AT1) are image heat maps that obtain gradients from VGG16, and the second attention (AT2) are object-labeled images.}
\label{fig_aoa_2}
\end{figure}

\vspace{5pt}
\textbf{Misaligned Image}. Objects appear at random positions in non-centrally aligned images, making it challenging to crop them for attention. To address this, we use Grad-CAM-based attention (As shown in Figure \ref{fig_aoa_2}). 

We feed images into a pre-trained VGG16 model \cite{VGG}, which consists of 13 convolutional layers within 5 blocks used for ImageNet classification. To crop the misaligned attention, we extract gradient values from the last convolutional layer and generate image heat maps. We refer to these heat maps as the first attention (AT1). Next, we label the objects on images to obtain object-labeled maps, which we refer to as the next attention (AT2).

To further improve the performance of our method, we incorporate both LPIPS and Gram-CAM into the VGG16 model. LPIPS scores are obtained by forward features, while Gram-CAM values are obtained through back feature gradients. We extract image features from the final convolutional layer of each block in VGG16 and compute the feature similarity to obtain LPIPS scores.

In addition to LPIPS, MSE, and COS, we add SSIM \cite{SSIM} to measure image similarity directly, as described in \cite{SSIM}. SSIM yields more informative gradients based on the previous loss functions, particularly in low-contrast areas \cite{understand_ssim}. Incorporating SSIM into the loss function can thus enhance the performance of GAN inversion. The image loss function is therefore based on a combination of these diverse similarity metrics.

\begin{equation}
\mathcal{L}_{IMG} = \alpha \mathcal{L}_{MSE} + \beta \mathcal{L}_{COS} + \gamma \mathcal{L}_{LPIPS} + \delta \mathcal{L}_{SSIM}.
\label{eq6}
\end{equation}

Where $\alpha, \beta, \gamma, \delta$ are hyper parameters that we set empirically to: $\alpha = 5$, $\beta = 3 $, $\gamma = 2$, $\delta = 1$. 

\subsection{Fine-tuning Pre-trained GAN Encoder}

Fine-tuned GAN encoders can find latent vectors that faithfully reconstruct real images. Fine-tuning improves the capacity of encoders to represent real image features that differ significantly from the pre-training datasets. This enhancement elevates the performance of pre-trained GANs in real image reconstruction tasks \cite{Ptuning}.

In the fine-tuning process in shallow GANs, our objective is to obtain latent vectors that represent real image inversions, e.g., ($z \simeq z'$) in shallow GANs or ($w \simeq w'$) in StyleGANs. Synthesized images ($\mathbf{x}$, $\mathbf{x'}$) should be replaced with real images, ($\mathbf{y}$, $\mathbf{y}'$). Therefore, the loss function evaluates two components: image ($IMG$) and latent space ($LS$).

\vspace{5pt}
\textbf{Latent Regularization}. We measure latent similarity using MSE and COS and utilize the following loss function to regularize inverted latent vectors:

\begin{equation}
\begin{aligned}
\mathcal{L}_{LS}(\mathbf{z} , \mathbf{z}') &= \alpha \mathcal{L}_{MSE}(\mathbf{z} ,\mathbf{z}') +  \beta \mathcal{L}_{COS}(\mathbf{z} , \mathbf{z}') \\
&= \alpha(\frac{||\mathbf{z}-\mathbf{z}'||_{2}}{n}) + \beta(1-\frac{\mathbf{z} \cdot {\mathbf{z}'}}{||\mathbf{z}||_2 \times ||\mathbf{z}'||_2}).
\label{eq_mtv}
\end{aligned}
\end{equation}

To fine-tune the encoder in shallow GANs, we evaluate the similarity in the two vectors: $(\mathbf{y}, \mathbf{y}')$ and $(\mathbf{z}, \mathbf{z}')$. However, small sizes of latent vectors struggle to represent HQ images. Therefore, HQ images are represented using layer-wise latent vectors, e.g., StyleGANs.

For StyleGAN inversion, prior works \cite{Style-GAN,Big-GAN} have shown that incorporating additional layer-wise and multi-type latent vectors can improve the representation of synthesized images. Hence, we introduced other types of latent vectors to jointly represent synthesized images. 

One of these is the constant vectors $\mathbf{z}_{c}$. The first block of the StyleGAN generator takes $\mathbf{z}_c$ as input, where the size of $\mathbf{z}_c$ is larger than the general latent vector $\mathbf{z}$. 

Another type of latent vector is the layer-level noise vector $\mathbf{z}_{n}$, which is used as a set of learnable parameters for each convolutional layer. $\mathbf{z}_{n}$ can control diverse local features for slightly modifying the peripheral areas of images, such as the local hair diversification in human faces.

We designed the StyleGAN encoder ($E$) that outputs $\mathbf{z}'_{c}$ and attaches learnable $\mathbf{z}'_{n}$ to convolutional layers. This incremental step ensures symmetric input and output between $E$ and $G$. Finally, $E$ outputs $\mathbf{w}'$, constant $\mathbf{z}_c'$, and upgrade learnable parameter $\mathbf{z}_n'$. We then evaluate the similarity between latent vectors via the encoder output. Considering two pairs of latent vectors ($\mathbf{w}$, $\mathbf{w}'$) and ($\mathbf{z}_c$, $\mathbf{z}_c'$), we regularize the latent vectors with the following loss function:

\begin{equation}
\mathcal{L}_{LS}(\mathbf{z},\mathbf{z}') \rightarrow \alpha \mathcal{L}_{LS}(\mathbf{w}, \mathbf{w}') + \beta \mathcal{L}_{LS}(\mathbf{z}_{c}, \mathbf{z}'_{c}) .
\label{eq1}
\end{equation}
Where $\alpha =1 $ and $\beta=1$ are hyper parameters. Here, the whole loss function for fine-tuning encoders is briefly summarized as (hyper parameters $\epsilon = 0.01$):

\begin{equation}
\mathcal{L}_{E} = \mathcal{L}_{IMG} + \epsilon \mathcal{L}_{LS}. 
\label{eq7}
\end{equation}

\vspace{5pt}
\section{Experiment}
\subsection{Setup}

\vspace{5pt}
\textbf{Device}. Our experiments were performed on a Nvidia Tesla V100-SXM3 (32GB) GPU. To invert real images, each image undergoes optimization within 3,500 iterations, approximately 2.08 iterations per second, and consumes about 7,670 MB of memory for the max resolutions (a 1024$\times$1024 real face image).

\vspace{5pt}
\textbf{Hyper-parameters}. We trained encoders for 7 epochs.  We used the Adam optimizer \cite{adam} with a learning rate of $0.0015$ and decayed $\beta$ values of $\beta_{1}=0$ and $\beta_{2}=0.99$. To maximize the CUDA memory usage during training, we set different hyper-parameters for varying image resolutions: batch sizes ($\mathbf{B}$), initial hidden features ($\mathbf{F}$), layer blocks ($\mathbf{L}$). For the $256 \times 256$ resolution: ($\mathbf{B} = 8$, $\mathbf{F}= 64 \times 64$, $\mathbf{F}= 7$);  for the $512 \times 512$ resolution: ($\mathbf{B} = 4$, $\mathbf{F} = 32 \times 32$, $\mathbf{L} = 8$); for the $1024 \times 1024$ resolution: ($\mathbf{B} = 2$, $\mathbf{F} = 16 \times 16$, $\mathbf{L} = 9$).

\vspace{5pt}
\textbf{Learning-parameters}. For most shallow GANs (e.g., PGGAN), the learning parameters are limited to the first layer input, i.e., $\mathbf{z'}$. For conditional GANs like BigGAN, we learned both the label vector $\mathbf{c'}$ and $\mathbf{z'}$ to ensure the reconstructed image corresponds to the correct label. For layer-wise representation GANs, such as StyleGANs, we learned the layer-wise vectors, i.e., $\mathbf{w'}$. To get the initial $\mathbf{w}$ for learning $\mathbf{w'}$, we used the general latent vector $\mathbf{z} \in \mathbb{R}^{512}$, which inputs into the pre-trained mapping model $M$ to obtain layer-wise $\mathbf{w} \in \mathbb{R}^{512 \times layers}$.

\vspace{5pt}
\textbf{Datasets}. In the pre-training step, we used pre-trained generators to synthesize 30,000 samples each, sourced from CelebA-HQ \cite{CelebAMask-HQ} and LSUN \cite{lsun} (using PGGAN and StyleGANs), as well as ImageNet (using BigGAN). We performed 7-fold cross-validation by randomly allocating 15\% samples (4,500) for each fold. In the fine-tuning step, we used 1,500 real images and implement 10-fold cross-validation by randomly allocating 10\% samples (150) for each fold.

\vspace{5pt}
\textbf{Evaluation Metrics}. We used PSNR, SSIM, and LPIPS to evaluate the similarity between original and reconstructed images for GAN inversion performance. To assess the quality of the GAN-generated inversion, we added FID \cite{fid}. 

\begin{table}[htbp]
\centering
\caption{Standard Deviations of Evaluation Methods (K = Folds, N = Samples)}
\label{sd}
\renewcommand\arraystretch{1.3} 
\setlength{\tabcolsep}{4pt} 
\begin{tabular}{|c|c|c|c|c|}
\hline \hline
                          & PSNR (dB)     & SSIM ($\%$)    &  LPIPS ($\%$)   & FID (Distance) \\
                          \hline
Pre-Training (K = 7, N = 4,500 ) & $0.81$ & $1.31\%$ & $1.38\%$ & $2.47$ \\
\hline
Fine-tuning (K = 10, N= 1,500)   & $1.54$ & $3.10\%$ & $2.92\%$ & $5.62$ \\  
\hline \hline
\end{tabular}
\end{table}

We reported the standard deviations of the cross-validation results for these evaluation methods in Table \ref{sd}. The results show that the standard deviations are greater in the fine-tuning step. We believe this issue arises because the data size in the fine-tuning step is smaller (4,500) than that in the pre-training step (30,000). Additionally, the features of real images differ more than those of synthesized images, leading to a greater standard deviation during fine-tuning.

\vspace{5pt}
\textbf{Comparisons}.  We compared synthesized images by famous GANs with those reconstructions, including PGGAN \cite{PGAN}, StyleGANs \cite{Style-GAN,StyleGAN2}, and BigGAN \cite{Big-GAN}, to evaluate the GAN inversion performance for the pre-trained GANs. We also compared our approach with existing state-of-the-art methods of GAN inversion, including RiDDLE \cite{RiDDLE}, E2Style \cite{E2Style}, pSp \cite{pSp}, In-Domain \cite{InDomainG}, ALAE \cite{ALAE} on StyleGANs (Fig. \ref{fig-compare-baseline}), and BigBiGAN \cite{BigBiGAN} on BigGAN (Fig. \ref{biggan-inversion}).

\vspace{5pt}
\textbf{Ablation Studies}. We conducted several ablation studies to investigate the impact of different training strategies (Fig. \ref{fig_ab_s1_s2}), cropped aligned attentions (Fig. \ref{fig_ab_s2}), misaligned attentions (Fig. \ref{fig_algin_misalign})  and the fine-tuning use of SSIM (Fig. \ref{ssim}). In addition, we compared the performance of our method with InterfaceGAN \cite{InterfaceGAN} (Fig. \ref{Real-img-processing})  on real image inversion and editing tasks. The ablation studies indicated the optimal settings for our method and allowed insight into the underlying factors affecting its performance.  

As shown in Table \ref{table-overview}, we summarize the subsequent experimental tables, including both baseline comparisons and ablation experiments.

\begin{table}[htbp]
\caption{Summary of Experimental Tables}
\label{table-overview}
\centering
\renewcommand\arraystretch{1.3} 
\setlength{\tabcolsep}{7pt} 
\begin{tabular}{|c|cc|cl|}
\hline \hline
\multirow{2}{*}{Table ID} &
  \multicolumn{2}{c|}{Experiment} &
  \multicolumn{2}{c|}{\multirow{2}{*}{Content}} \\ \cline{2-3}
        & \multicolumn{1}{c|}{Baseline comparison} & Ablation study & \multicolumn{2}{c|}{}         \\ \hline
Table \ref{tab-relatedworks} & \multicolumn{1}{c|}{$\checkmark$}        &                & \multicolumn{2}{c|}{StyleGANs} \\ \hline
Table \ref{bigGAN_inversion} &
  \multicolumn{1}{c|}{\begin{tabular}[c]{@{}c@{}}$\checkmark$\end{tabular}} &
   &
  \multicolumn{2}{c|}{BigGAN} \\ \hline
Table \ref{tab-3-scale-img} &
  \multicolumn{1}{c|}{} &
  \begin{tabular}[c]{@{}c@{}}$\checkmark$\end{tabular} &
  \multicolumn{2}{c|}{Cropping image} \\ \hline
Table \ref{ab_ssim} &
  \multicolumn{1}{c|}{} &
  \begin{tabular}[c]{@{}c@{}}$\checkmark$\end{tabular} &
  \multicolumn{2}{c|}{Fine-tuining(SSIM)} \\ \hline
Table \ref{image_edit}  &
  \multicolumn{1}{c|}{\begin{tabular}[c]{@{}c@{}}$\checkmark$\end{tabular}} &
   &
  \multicolumn{2}{c|}{Face editing} \\ \hline \hline
\end{tabular}
\end{table}

\subsection{Comparison of GAN Inversion}

\textbf{Pre-trained GAN Inversion}. Our method is capable of reconstructing GAN-synthesized images and repairing local areas that may have produced crumbled or unreasonable objects. We attribute this improvement to the latent representation learned by our pre-trained encoder.

In the case of BigGAN, the reconstructions are nearly indistinguishable from the synthesized images. We attribute this success to the inclusion of layer-wise label vectors ($\mathbf{c}$). In the case of PGGAN and StyleGAN1, where some results are imperfect, our method can reconstruct images better than the synthesized images. In the case of StyleGAN2, our method keeps the same performance compared to StyleGAN2. The results are presented in the appendix (see Fig. \ref{fig-compare-BigGAN} and Fig. \ref{fig-compare-GANs}).

\begin{figure*}[h!]
\centering
\includegraphics[width=\linewidth]{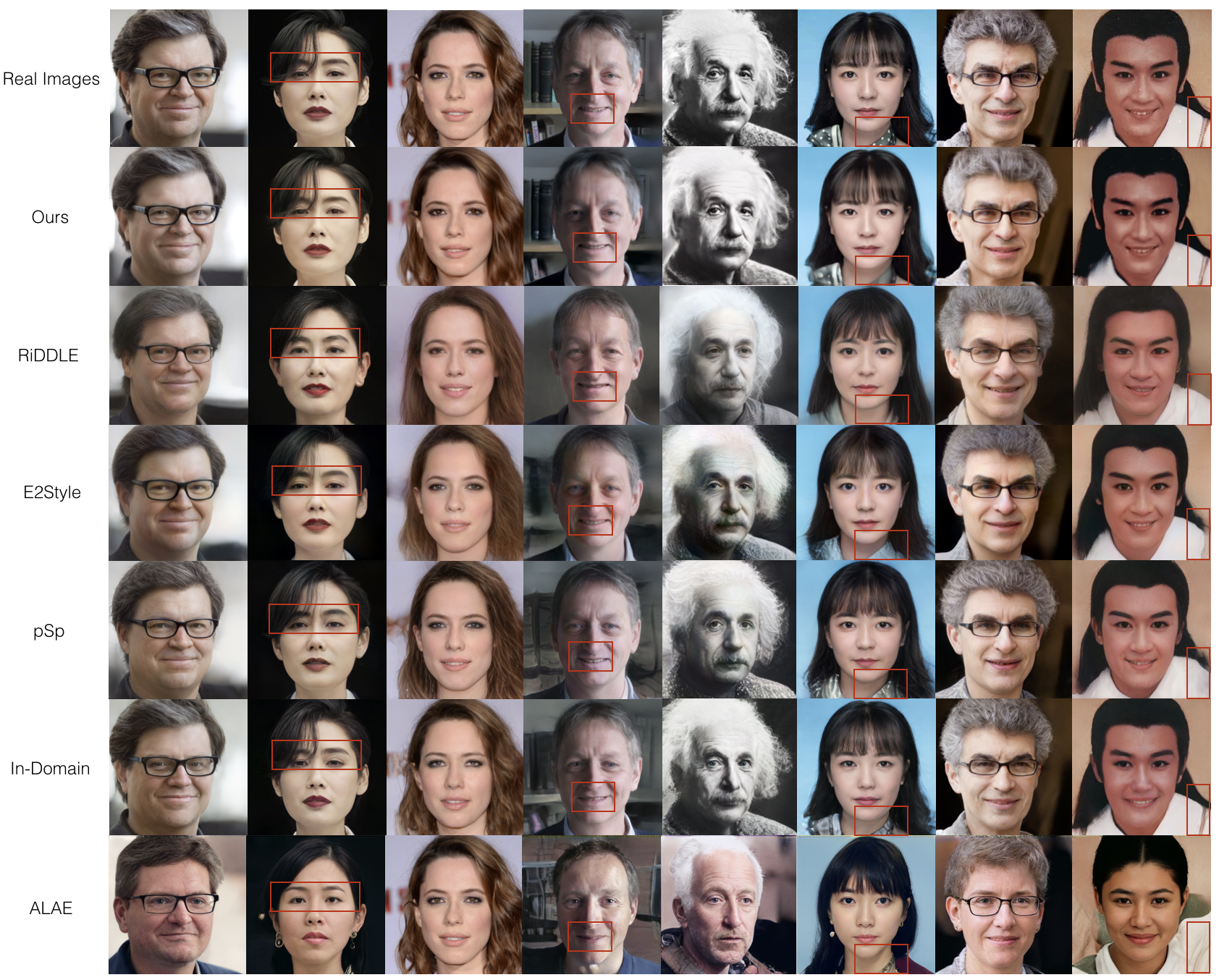} 
\caption{Qualitative comparison of our method with RiDDLE \cite{RiDDLE}, E2Style \cite{E2Style}, pSp \cite{pSp}, InDomainG \cite{InDomainG}, and ALAE \cite{ALAE} on real face inversions. Red windows in the images alternately highlight the different details. Our method outperforms in red  windows (zoom-in images for better visual comparison). }
\label{fig-compare-baseline}
\end{figure*}

\textbf{Baseline Comparisons}. On the task of real face inversion, we compared our method with state-of-the-art methods, i.e., RiDDLE \cite{RiDDLE}, E2Style \cite{E2Style}, pSp \cite{pSp}, InDomainG \cite{InDomainG}, and ALAE \cite{ALAE}. Except for ALAE, which can hardly reconstruct the original face, other methods are more successful, but our method performs better in small details. We present the visual comparison in Fig. \ref{fig-compare-baseline}.

ALAE, pSp, and E2Style inverted images directly from pre-trained $E$. Similar to  \cite{img2stylegan}, In-Domain optimized $\mathbf{w}$ based on its pre-trained encoder. In contrast, we fine-tuned $E$ based on the pre-trained $E$, then let $E$ invert real images to ideal $\mathbf{w}$. 

We quantitatively evaluated the image similarity on synthesized ($\mathbf{x}$ vs. $\mathbf{x}'$) and real images ($\mathbf{y}$ vs. $\mathbf{y}'$). Here, we did not compare ALAE in real images because ALAE cannot effectively reconstruct real images. We report results in Table \ref{tab-relatedworks}. 

\begin{table}[htbp]
\small
\centering
\caption{Quantitative comparison of StyleGAN inversion methods on synthesized and real images.}
\label{tab-relatedworks}
\renewcommand\arraystretch{1.3} 
\setlength{\tabcolsep}{3.3pt} 
\begin{tabular}{|c|c|c|c|c|}
\hline \hline
$\mathbf{x}$ vs. $\mathbf{x'}$ & PSNR$\uparrow$ & SSIM$\uparrow$  & LPIPS$\downarrow$ & FID$\downarrow$ \\ 
\hline 
ALAE\cite{ALAE} & 15.68 & 0.497 & 0.485 & 114.48\\
In-domain\cite{InDomainG}  & 19.53 & 0.584  & 0.422  & 102.05 \\
pSp\cite{pSp} &  21.53 & 0.640  & 0.398  & \textbf{77.09} \\
E2Style\cite{E2Style} & \textbf{22.03} & \textbf{0.667} & 0.394 & 90.37 \\
Ours  & 20.45 & 0.630  & \textbf{0.368}  & 81.63\\
\hline \hline
$\mathbf{y}$ vs. $\mathbf{y'}$  & \quad & \quad & &\\ 
\hline 
In-domain\cite{InDomainG} & 23.45 & 0.672  & 0.367  & 59.12\\
pSp\cite{pSp} & 21.38 & 0.649  & 0.416  & 96.48\\ 
E2Style\cite{E2Style} & 22.03 & 0.683 & 0.394 & 93.66 \\
RiDDLE\cite{RiDDLE} & 23.62 & 0.724  & 0.371 & 86.19\\
Ours & \textbf{25.92} & \textbf{0.767}  & \textbf{0.293}  & \textbf{44.52}\\
\hline \hline

\end{tabular}
\end{table}

RiDDLE, E2Style, and pSp provide better results for GAN-synthesized images and are also good for real images. For real image inversion, our method outperforms other methods by fine-tuning pre-trained $E$.

We also compared our method to BigBiGAN \cite{BigBiGAN} for BigGAN \cite{Big-GAN} on the ImageNet dataset. While BigGAN has lower performance than StyleGANs, it is currently the only valid ImageNet-compatible method. The results demonstrate that our method outperforms BigBiGAN within the constraints of the ImageNet environment.

\begin{figure}[h!]
\centering 
\includegraphics[width=0.55\linewidth]{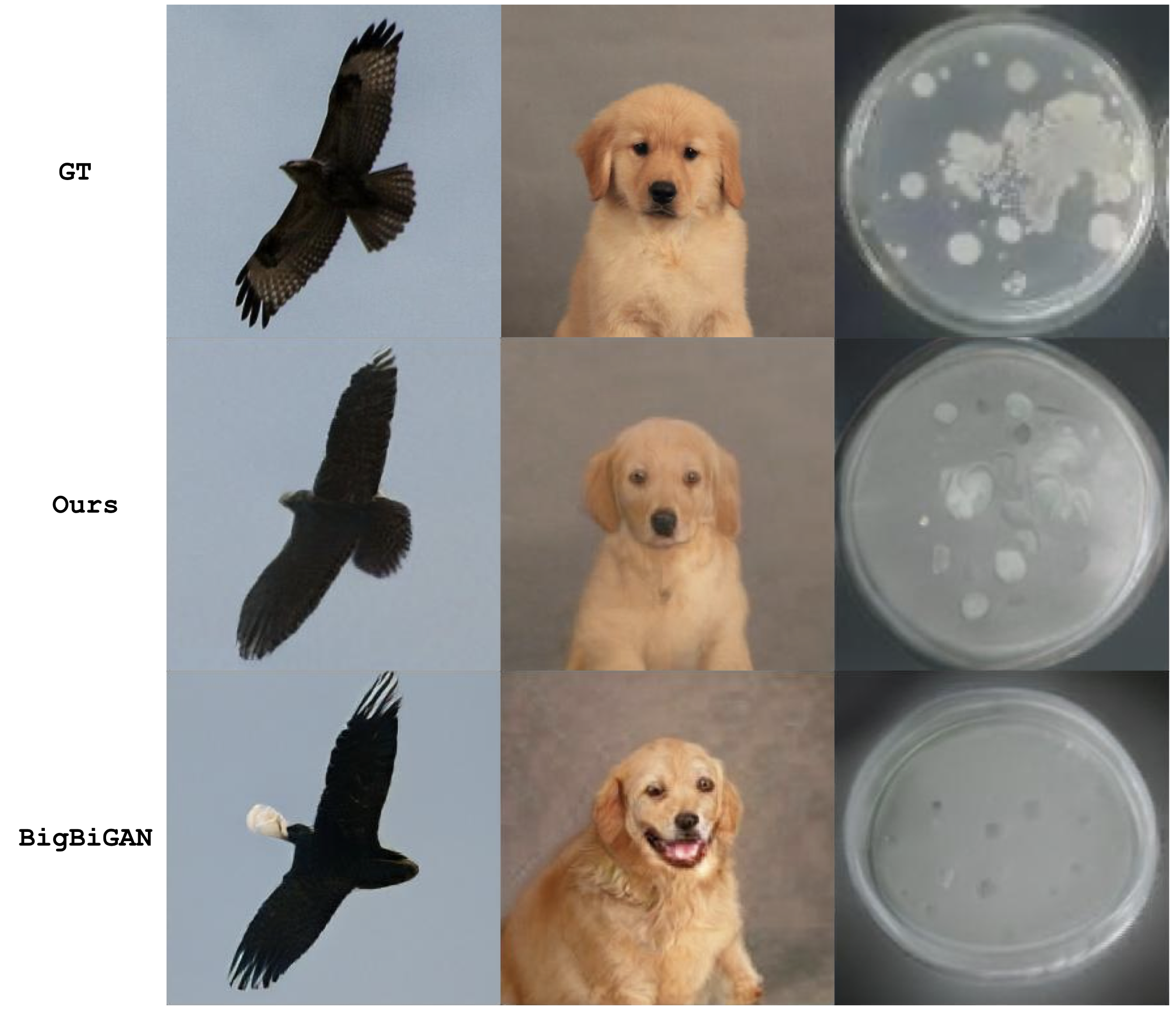} 
\caption{Qualitative comparison of our method and BigBiGAN \cite{BigBiGAN} for inverting 256x256 ImageNet images. Three labels used for the demonstration are bald eagle, golden retriever, and petri dish. Our method demonstrated superior results to BigBiGAN in GAN inversion.}
\label{biggan-inversion}
\end{figure}

BigBiGAN is a BiGAN-inspired method \cite{Bi_GAN} to invert images generated by BigGAN. However, BigBiGAN was unable to invert most real images due to pattern collapse during BigGAN inversion. We present three representative ImageNet cases in Fig. \ref{biggan-inversion} and report the qualitative evaluation in Table \ref{bigGAN_inversion}.

\begin{table}[h!]
\small
\centering
\caption{Quantitative comparison of BigGAN inversion on ImageNet. Our method vs. BigBiGAN.}
\label{bigGAN_inversion}
\renewcommand\arraystretch{1.3} 
\setlength{\tabcolsep}{3.7pt} 
\begin{tabular}{|l|c|c|c|c|}
\hline \hline

& \multicolumn{1}{l|}{PSNR$\uparrow$} & SSIM$\uparrow$ & LPIPS$\downarrow$ & FID$\downarrow$ \\ \hline

BigBiGAN \cite{BigBiGAN}  & 25.14                               & 0.750          & 0.348             & 106.58          \\ \hline
Ours              & \textbf{26.63}                      & \textbf{0.761} & \textbf{0.329}    & \textbf{89.26}  \\ \hline \hline
\end{tabular}
\end{table}

\subsection{Ablation Study}

\textbf{Training Strategy}. There are two pre-training strategies for GAN encoders, as shown below: 

\vspace{5pt}
Strategy 1: we set $\mu_1$ = 1 and $\mu_2$ = 1 (as shown in Eq. \ref{eq6}). Here, we use two cropped attentions to improve the performance. To conserve GPU memory, we removed the pixel gradients during backward propagation. 

\vspace{5pt}
Strategy 2: we set $\mu_1$ = 5 and $\mu_2$ = 9, and added a fused cropped operation in each block, which is similar to PGGAN \cite{PGAN}, yield slightly better performances on attention regions with larger values of $(\mu_1, \mu_2)$. 

\begin{figure}[htbp]
\centering 
\includegraphics[width=0.85\linewidth]{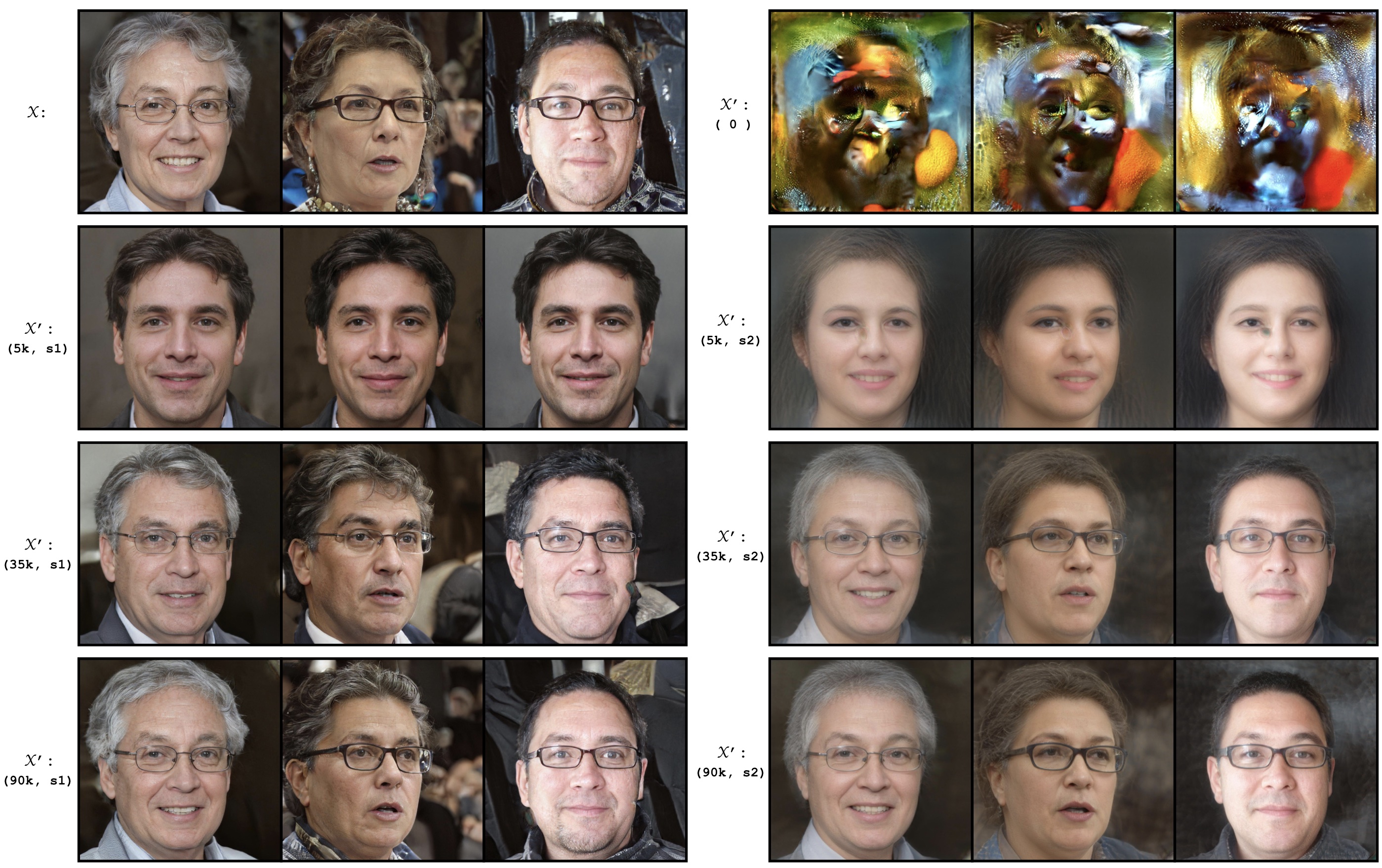} 
\caption{The two strategies differ in their treatment of the last pixel gradient. Strategy 1 (S1) trains the encoder without the last pixel gradients, while Strategy 2 (S2) trains the encoder with pixel gradients.}
\label{fig_ab_s1_s2}
\end{figure}

\vspace{5pt}
Although Strategy 2 provides more faithful image reconstructions with preserved attention gradients, the overhead cost of this approach should be considered, particularly concerning memory and computation requirements. Specifically, the training process for Strategy 2 consumes approximately $20\%$ more gradient memory in StyleGAN2 FFHQ, which limits its scalability for HQ images. Additionally, Strategy 1 optimizes encoders from high-level features (non-pixel). In contrast, Strategy 2 trains encoders from pixel level at random noise, resulting in a gradient trajectory that transitions from noise to blurred images. Figure \ref{fig_ab_s1_s2} visually compares these two strategies.

\vspace{5pt}
\textbf{Cropping Attentions}. To overcome the challenge of reconstructing HQ images for GAN inversion, we evaluated the differences of cropping attentions by  reconstructing 1024$\times$1024 faces ($x$) at the 2nd epoch. For comparison, we trained $E$ via different cropped attentions and evaluated the inversion performances. 

\begin{figure}[htbp]
\centering 
\includegraphics[width=0.8\linewidth]{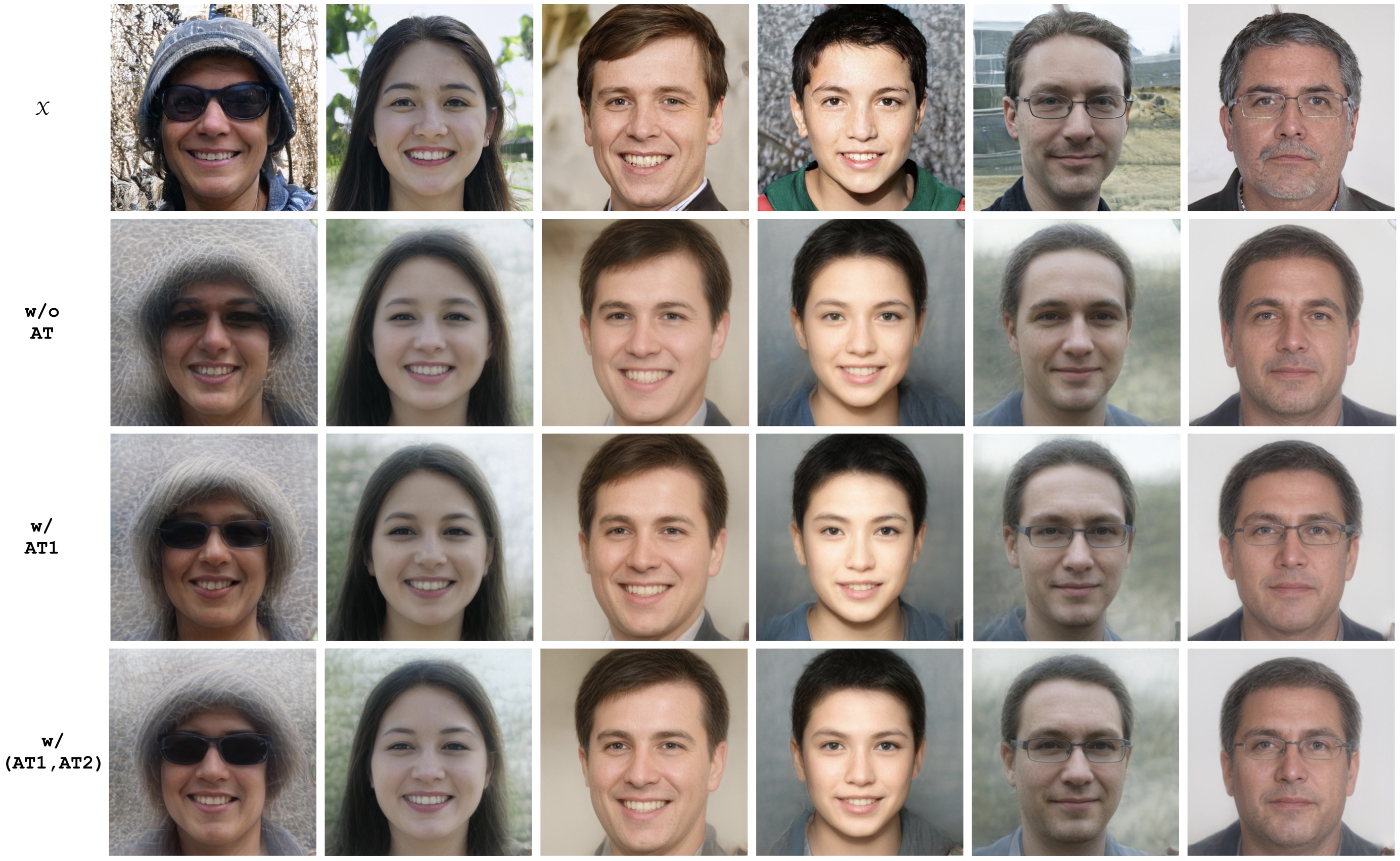} 
\caption{Visual comparison of w/o center-aligned attentions using Strategy 2. The 1st row shows generated faces by StyleGAN2 (FFHQ, config F). The 2nd--4th rows show reconstructed images with different attentions (AT1 and AT2). With the addition of different cropped attentions, the visual performance is gradually improved.}
\label{fig_ab_s2}
\end{figure}

We evaluated different attentions (none, one, or two) on image reconstruction performance, as shown in Fig. \ref{fig_ab_s2}. Adding AT1 improved the similarity of the main reconstructive area, and adding AT2 further improved similarity, especially in key image areas. The best performance was observed with the progressive addition of AT1 and AT2. We report the quantitative evaluations in Table \ref{tab-3-scale-img}.

\begin{table}[h!]
\small
\centering
\caption{Quantitative comparison for the ablation study of adding cropped attentions.}
\label{tab-3-scale-img}
\renewcommand\arraystretch{1.3} 
\setlength{\tabcolsep}{3.3pt} 
\begin{tabular}{|cc|c|c|c|c|}
\hline \hline
\multicolumn{2}{|c|}{\multirow{2}{*}{Reconstructions}} &
  \multirow{2}{*}{PSNR$\uparrow$} &
  \multirow{2}{*}{SSIM$\uparrow$} &
  \multirow{2}{*}{LPIPS$\downarrow$} &
  \multirow{2}{*}{FID$\downarrow$} \\
\multicolumn{2}{|c|}{}                                             &     &      &     &        \\ \hline \hline
\multicolumn{1}{|c|}{\multirow{3}{*}{$\mathbf{x'}$}}& w/o AT & \textbf{18.97} & \textbf{0.699} & 0.411 & 165.45 \\ \cline{2-6} 
\multicolumn{1}{|c|}{}                              & w/ AT1 & 18.82	 & 0.698 & \textbf{0.409} & \textbf{149.34}    \\ \cline{2-6} 
\multicolumn{1}{|c|}{}                              
& w/ (AT1, AT2) & 18.75  & 0.697   & 0.412  & 152.80   
\\ \hline \hline
\multicolumn{1}{|c|}{\multirow{3}{*}{$\mathbf{x_1'}$}} 
& w/o ATs   &  19.65  & 0.685 & 0.404   & 135.34      \\ \cline{2-6} 
\multicolumn{1}{|c|}{}                              
&  w/ AT1   &  \textbf{19.76}  & 0.686 & \textbf{0.401}   & \textbf{127.25}        \\ \cline{2-6} 
\multicolumn{1}{|c|}{}                              
& w/ (AT1, AT2) & 19.71  & \textbf{0.688}   & 0.402  & 136.48 
\\ \hline \hline
\multicolumn{1}{|c|}{\multirow{3}{*}{$\mathbf{x_2'}$}} 
&  w/o AT   &  20.79   & 0.701    & 0.386     & 101.63      \\ \cline{2-6} 
\multicolumn{1}{|c|}{}                              
& w/ AT1    &  20.72   & 0.699    & 0.381     & 87.23      \\ \cline{2-6} 
\multicolumn{1}{|c|}{}                              
& w/ (AT1, AT2) & \textbf{21.23}  & \textbf{0.712}  & \textbf{0.377}  & \textbf{86.09} 
\\ \hline \hline
\end{tabular}
\end{table}

\begin{figure}[htbp]
\centering
\includegraphics[width=0.7\linewidth]{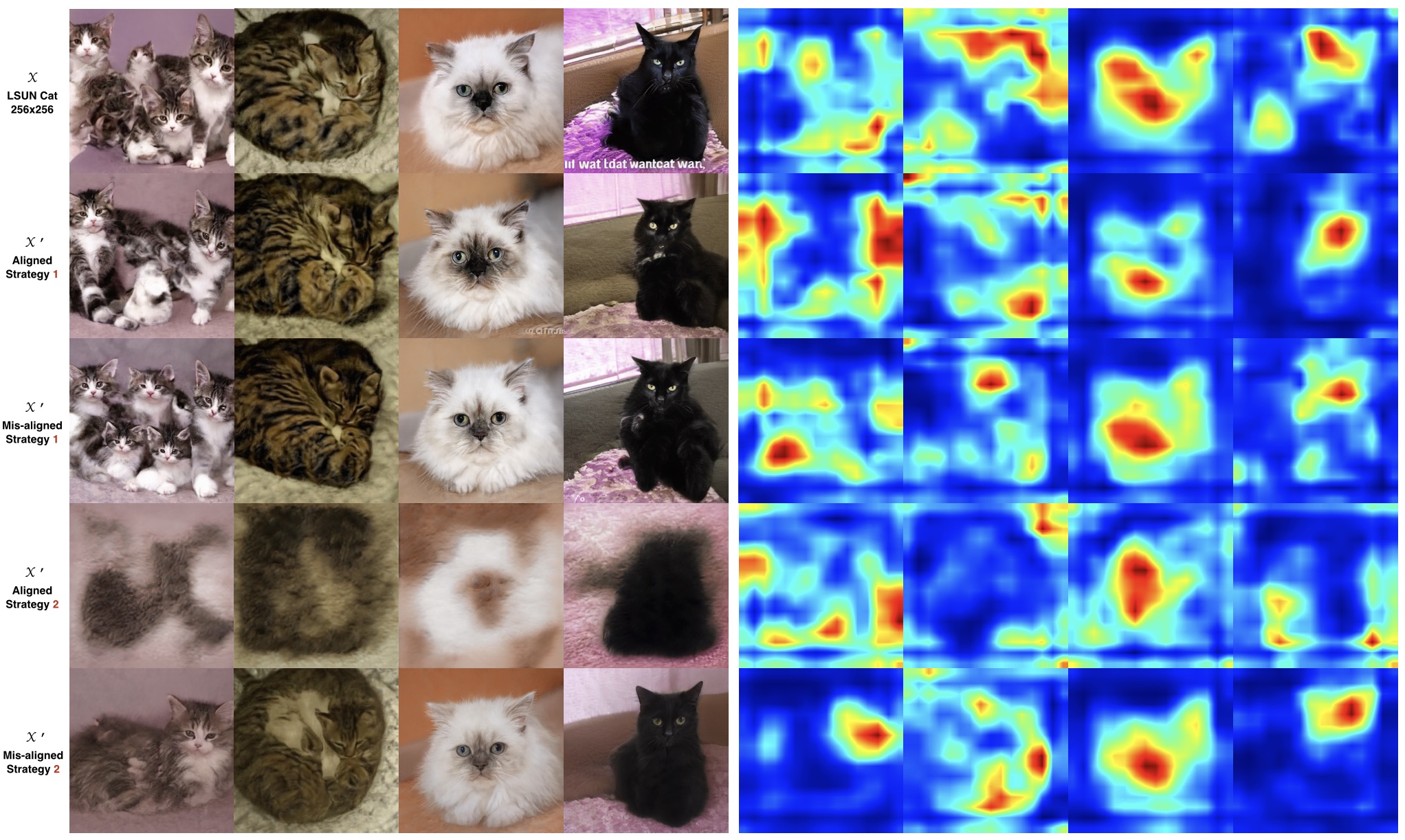} 
\caption{Visual comparison of center-aligned and Grad-CAM on misaligned images among 2 training strategies on StyleGAN2 LSUN cats (256$\times$256). In training strategy 1, Grad-CAM attentions outperformed center-aligned attentions (see rows 2 and 3). In strategy 2, training $E$ turns to mode collapse. In contrast, Gram-CAM attention produced reliable results using strategy 2 (last row).}
\label{fig_algin_misalign}
\end{figure}

\textbf{Misaligned Images}. To compare the effectiveness of center-aligned attentions and Grad-CAM attentions for misaligned images, we evaluated the dataset of LSUN cats, where cats are misaligned objects in images. As shown in Fig. \ref{fig_algin_misalign}, we evaluated two training strategies. Compared with center-aligned attention, Grad-CAM attention produced better results.

\begin{figure}[h!]
\centering
\includegraphics[width=0.55\linewidth]{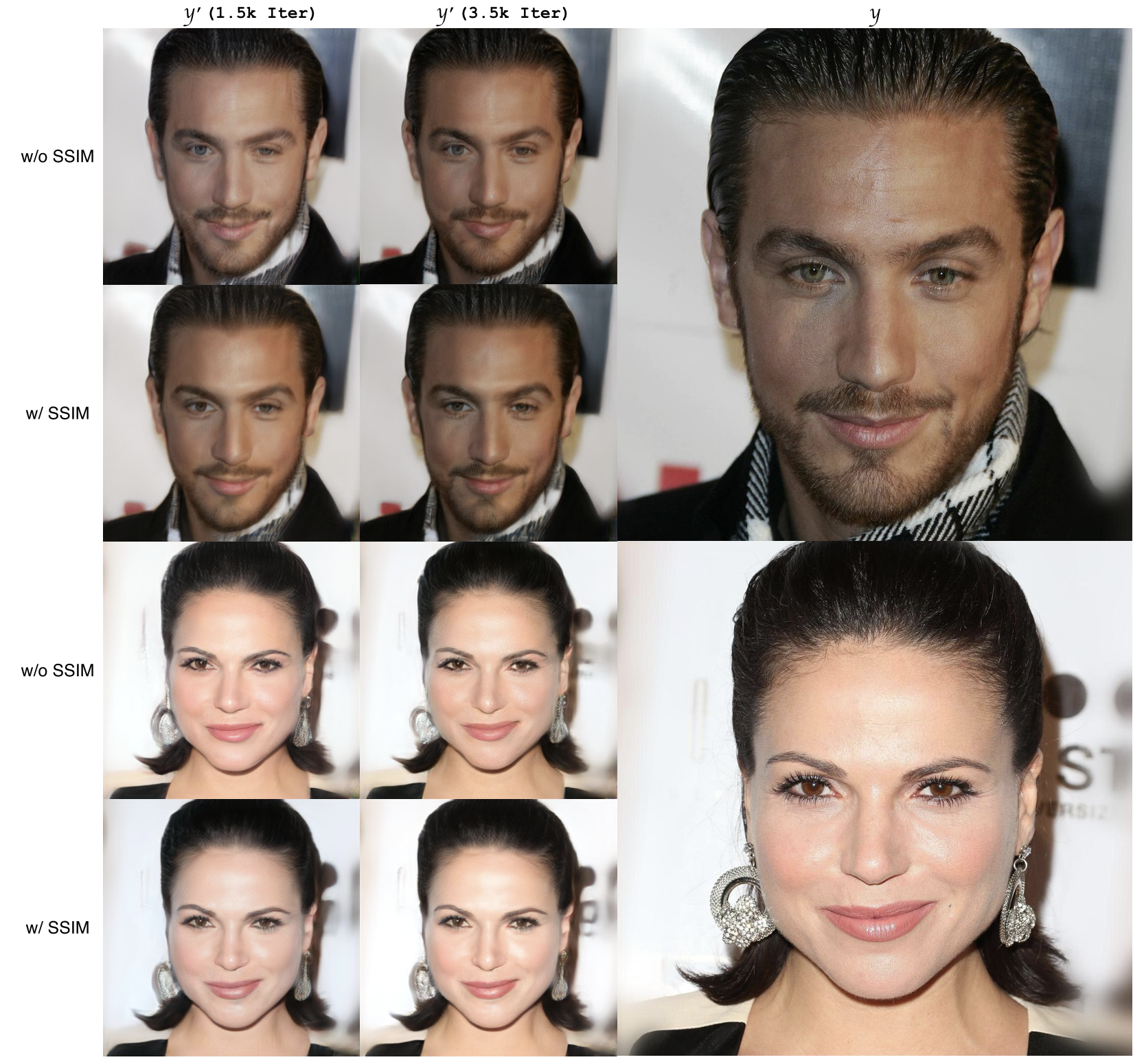} 
\caption{Visual comparison of w/o SSIM similarity on real face inversion. Adding SSIM improve the  performance of real face inversions.}
\label{ssim}
\end{figure}

\vspace{5pt}
\textbf{w/o SSIM}. Unlike previous works that use MSE, LPIPS, or face identity models to evaluate image similarity \cite{img2stylegan,InDomainG,pSp}, we include SSIM to the loss function to train and fine-tune encoders. Fig. \ref{ssim} shows the visual comparison and confirms that using SSIM leads to better inversions, which were evaluated at two different iterations. Table \ref{ab_ssim} shows the quantitative results for GAN inversion evaluated on the CelebA-HQ datasets. These results demonstrated that we improved performance with the addition of SSIM.

\begin{table}[htbp]
\small
\centering
\renewcommand\arraystretch{1.3} 
\setlength{\tabcolsep}{3.3pt} 
\caption{Quantitative comparison for the ablation study of w/o SSIM.}
\label{ab_ssim}
\begin{tabular}{|c|c|c|c|c|}
\hline \hline
\multicolumn{1}{|c|}{\multirow{2}{*}{1,500 iterations}} &
  \multicolumn{1}{c|}{\multirow{2}{*}{PSNR$\uparrow$}} &
  \multicolumn{1}{c|}{\multirow{2}{*}{SSIM$\uparrow$}} &
  \multicolumn{1}{c|}{\multirow{2}{*}{LPIPS$\downarrow$}} &
  \multirow{2}{*}{FID$\downarrow$} \\
\multicolumn{1}{|c|}{}         & \multicolumn{1}{c|}{}    & \multicolumn{1}{c|}{}    & \multicolumn{1}{c|}{}     &    \\ \hline
\multicolumn{1}{|c|}{w/o SSIM} & \multicolumn{1}{c|}{21.92} & \multicolumn{1}{c|}{0.632} & \multicolumn{1}{c|}{\textbf{0.395}} & \multicolumn{1}{c|}{78.10}  \\ \hline
\multicolumn{1}{|c|}{w/ SSIM}   & \multicolumn{1}{c|}{\textbf{23.07}} & \multicolumn{1}{c|}{\textbf{0.657}} & \multicolumn{1}{c|}{0.396} & \multicolumn{1}{c|}{\textbf{68.06}}   \\ \hline \hline
\multicolumn{5}{|l|}{3,500 iterations}                                                                                                                  \\ \hline
\multicolumn{1}{|c|}{w/o SSIM} & \multicolumn{1}{c|}{22.90}   & \multicolumn{1}{c|}{0.639}  & \multicolumn{1}{c|}{\textbf{0.386}}  & \multicolumn{1}{c|}{76.06}    \\ \hline
\multicolumn{1}{|c|}{w/ SSIM}   & \multicolumn{1}{c|}{\textbf{23.19}}  & \multicolumn{1}{c|}{\textbf{0.657}}  & \multicolumn{1}{c|}{0.393}   & \multicolumn{1}{c|}{\textbf{65.03}}     \\ \hline \hline
\end{tabular}
\end{table}

\begin{figure}[htbp]
\centering
\includegraphics[width=0.85\linewidth]{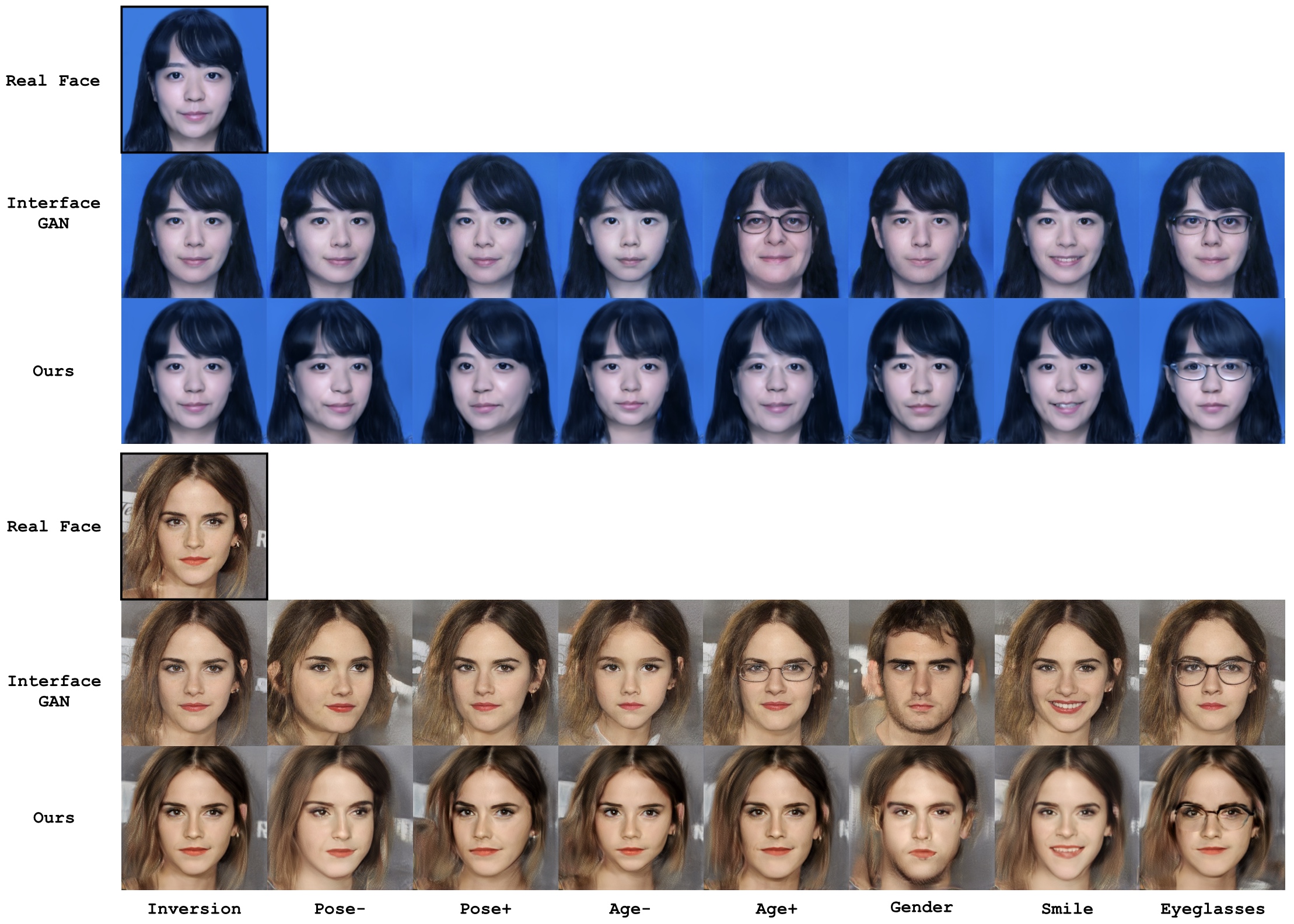} 
\caption{Qualitative comparison of face editing. We compare our method to InterfaceGAN \cite{InterfaceGAN} in terms of real face reconstruction and face editing using five learned latent directions. The latent directions for pose and age display both positive (+) and negative (-) results.}
\label{Real-img-processing}
\end{figure}

\subsection{Real Face Editing}
As shown in Fig. \ref{Real-img-processing}, real faces can be edited through learned latent directions using face attribute labels. In comparison with InterfaceGAN \cite{InterfaceGAN} and its inversion method \cite{InDomainG}, our approach improves GAN inversion and latent regularization to achieve state-of-the-art performance. 

To further evaluate the performance of our method with InterfaceGAN, we conducted a quantitative evaluation that compared both method results covering five face attribute labels (pose, age, gender, smile, and eyeglasses). We utilized \cite{RFM} to learn our label-based directions. The compared ground truth are real faces. The results are in Table \ref{image_edit}.

\begin{table}[htbp]
\small
\centering
\renewcommand\arraystretch{1.3} 
\setlength{\tabcolsep}{3.3pt} 
\caption{Quantitative Comparison of Face Editing (Our method vs. InterfaceGAN).}
\label{image_edit}
\begin{tabular}{|c|c|c|c|c|}
\hline \hline
\multicolumn{1}{|c|}{\multirow{2}{*}{}} &
  \multicolumn{1}{c|}{\multirow{2}{*}{PSNR$\uparrow$}} &
  \multicolumn{1}{c|}{\multirow{2}{*}{SSIM$\uparrow$}} &
  \multicolumn{1}{c|}{\multirow{2}{*}{LPIPS$\downarrow$}} &
  \multirow{2}{*}{FID$\downarrow$} \\
\multicolumn{1}{|c|}{}         & \multicolumn{1}{c|}{}    & \multicolumn{1}{c|}{}    & \multicolumn{1}{c|}{}     &    \\ \hline
\multicolumn{1}{|c|}{InterfaceGAN \cite{InterfaceGAN}} & \multicolumn{1}{c|}{18.28} & \multicolumn{1}{c|}{0.553} & \multicolumn{1}{c|}{0.465} & \multicolumn{1}{c|}{93.23}  \\ \hline
\multicolumn{1}{|c|}{Ours}   & \multicolumn{1}{c|}{\textbf{19.70}} & \multicolumn{1}{c|}{\textbf{0.561}} & \multicolumn{1}{c|}{\textbf{0.428}} & \multicolumn{1}{c|}{\textbf{87.59}}   \\ \hline \hline
\end{tabular}
\end{table}

\section{Limitation}

\begin{figure}[htbp]
\centering
\includegraphics[width=0.73\linewidth]{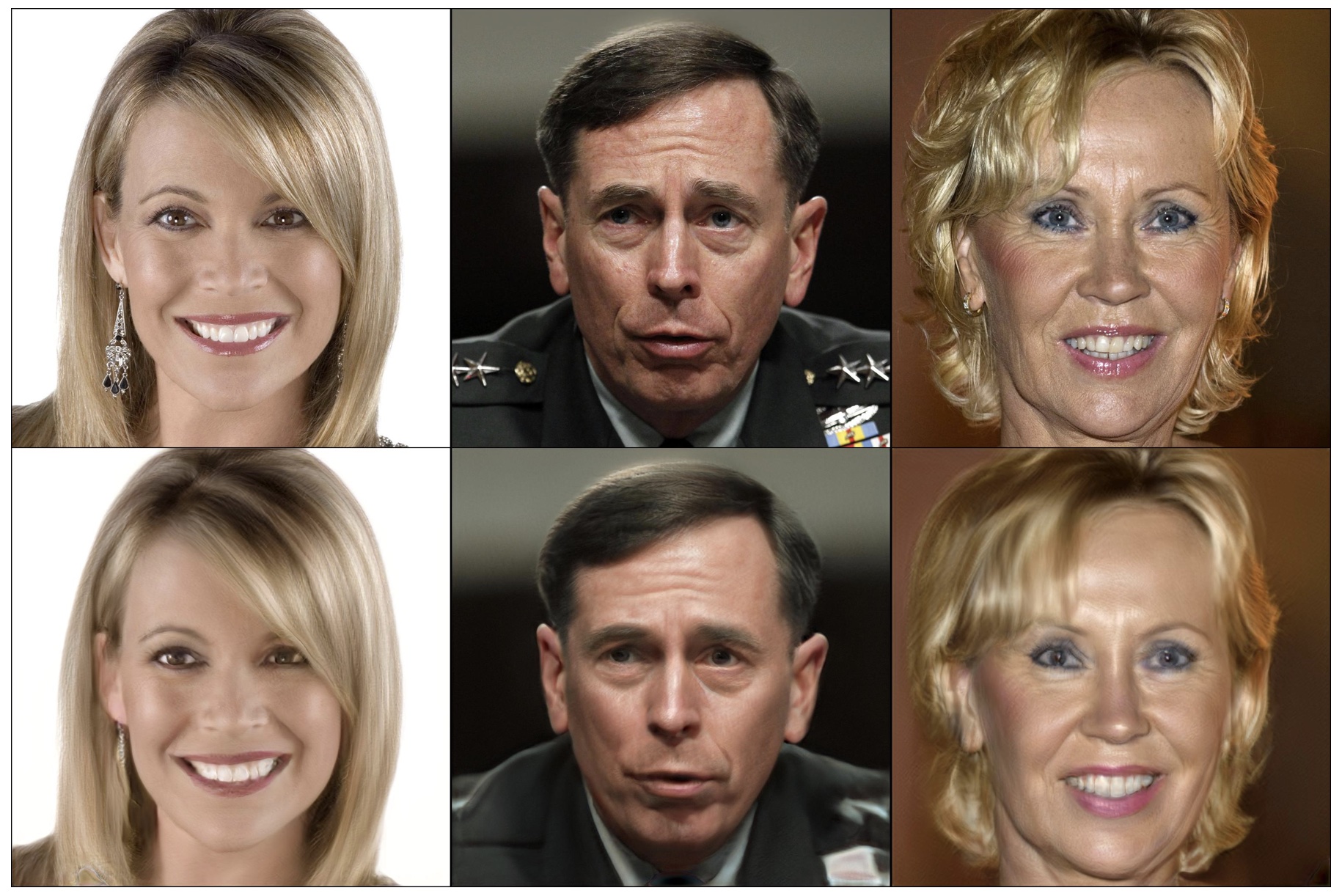} 
\caption{GAN inversion deficiency for high-quality real faces. This leads to the blurring of accessories such as the woman's one earring (in the first column), the man's badge (in the second column), and the woman's two earrings (in the third column).}
\label{realface_shortage}
\end{figure}

Current GAN inversion methods limit the quality of HQ real face reconstructions. As shown in Fig. \ref{realface_shortage}, existing methods struggle to accurately preserve small details in reconstructed faces, such as earrings and badges. We believe the current state-of-the-art GANs inadequately generate these features, resulting in unsuccessful reconstructions.

\section{Conclusion}

In this paper, we present a novel GAN inversion approach that facilitates the reconstruction of real images via fine-tuning pre-trained GAN encoders. Our approach was successfully implemented in various popular GANs, including PGGAN, BigGAN, and StyleGANs. Additionally, we improved image similarity evaluation in the loss function by using cropping attentions to highlight the objects of interest. In our future work, we plan to explore the learning of interpretable directions in latent space to enable real image editing.

\section*{Declaration of Competing Interest}
The authors declare that there are no known competing financial interests or personal relationships that could have appeared to influence the work reported in this paper.

\section*{Ethical and Informed Consent for Data Used}
Ethical and informed consent for data used in this paper was obtained from all authors.

\section*{Data Availability and Access}
All data included in this study are available upon request by contact with the authors.

\section*{Acknowledgements}
This work was supported by the Science and Technology Development Fund (FDCT) of Macau; Natural Science Foundation of Chongqing, China (CSTB2023NSCQ-LZX0068); Scientific Research Foundation of Chongqing University of Technology (2023ZDZ022), and Science and Technology Research Program of Chongqing Education Commission of China (Youth Project-KJQN202401159).

\section*{Appendix}
\label{Appendix}

\begin{figure*}[htbp]
\centering
\includegraphics[width=\linewidth]{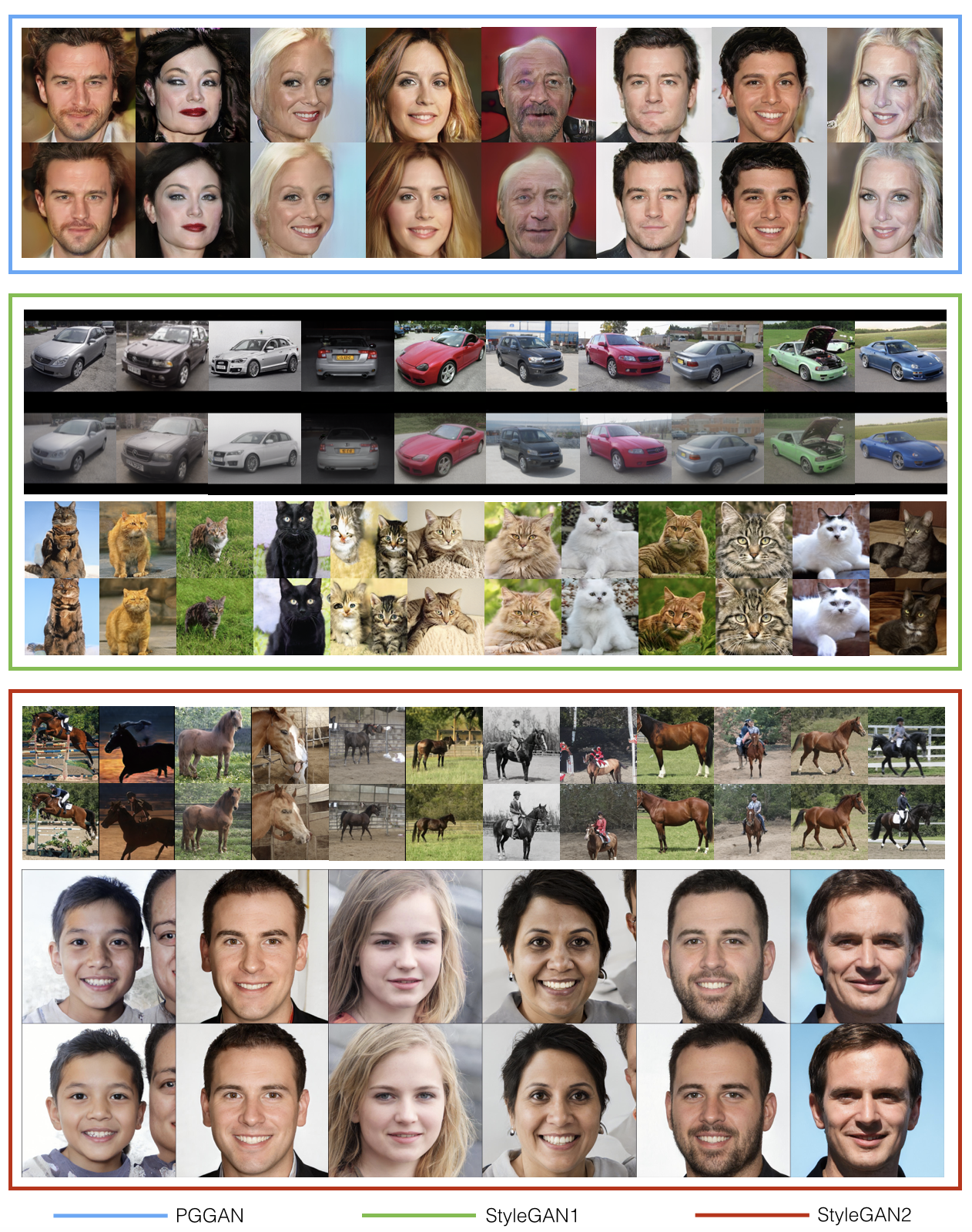} 
\caption{Visual comparison of GAN-synthesized images (upper row) with our reconstructions (lower row) from PGGAN (FFHQ 1024x1024 faces), StyleGAN1 (LSUN 512x512 cars and 256x256 cats), and StyleGAN2 (LSUN 256x256 horses and 1024x1024 faces). Our method successfully reproduces or enhances the synthesized images of these GANs}
\label{fig-compare-GANs}
\end{figure*}

As shown in Tables \ref{tab_model_1} and \ref{tab_model_2}, we report the different encoder architectures for inverting StyleGANs, PGGAN, and BigGAN. We also report more visual results of GAN inversion in Fig. \ref{fig-compare-GANs}.

\begin{table*}[htbp]
\tiny
\centering
\caption{Architectures for StyleGAN encoders.}
\label{tab_model_1}
\renewcommand\arraystretch{1.5} 
\setlength{\tabcolsep}{9.9pt} 
\newcommand{\tabincell}[2]{\begin{tabular}{@{}#1@{}}#2\end{tabular}} 
\begin{tabular}{|c|c|c|c|}
\hline
\multirow{2}{*}{\tabincell{c}{Block ID \\ Res.}} 
& \multicolumn{3}{c|}{StyleGAN Encoder ($E$)} \\ \cline{2-4} 
& 256$\times$256 & 512$\times$512 & 1024$\times$1024 \\ \hline \hline \hline
\multirow{2}{*}{\tabincell{c}{1\\ 4$\times$4}} 
& FC, IN, CONV(64,64,3), $\mathbf{z}'_{n}$, L-ReLU 
& FC, IN, CONV(32,32,3), $\mathbf{z}'_{n}$, L-ReLU 
& FC, IN, CONV(16,16,3), $\mathbf{z}'_{n}$, L-ReLU \\ \cline{2-4} 

& FC, IN, CONV(64,128,3), $\mathbf{z}'_{n}$, L-ReLU 
& FC, IN, CONV(32,64,3), $\mathbf{z}'_{n}$, L-ReLU 
& FC, IN, CONV(16,32,3), $\mathbf{z}'_{n}$, L-ReLU \\ \hline \hline

\multirow{2}{*}{\tabincell{c}{2\\ 8$\times$8}} 
& FC, IN, CONV(128,128,3), $\mathbf{z}'_{n}$, L-ReLU
& FC, IN, CONV(64,64,3), $\mathbf{z}'_{n}$, L-ReLU 
& FC, IN, CONV(32,32,3), $\mathbf{z}'_{n}$, L-ReLU \\ \cline{2-4} 
& FC, IN, CONV(128,256,3), $\mathbf{z}'_{n}$, L-ReLU 
& FC, IN, CONV(64,128,3), $\mathbf{z}'_{n}$, L-ReLU 
& FC, IN, CONV(32,64,3), $\mathbf{z}'_{n}$, L-ReLU \\ \hline \hline

\multirow{2}{*}{\tabincell{c}{3\\ 16$\times$16}} 
& FC, IN, CONV(256,256,3), $\mathbf{z}'_{n}$, L-ReLU 
& FC, IN, CONV(128,128,3), $\mathbf{z}'_{n}$, L-ReLU 
& FC, IN, CONV(64,64,3), $\mathbf{z}'_{n}$, L-ReLU \\ \cline{2-4} 

& FC, IN, CONV(256,512,3), $\mathbf{z}'_{n}$, L-ReLU 
& FC, IN, CONV(128,256,3), $\mathbf{z}'_{n}$, L-ReLU 
& FC, IN, CONV(64,128,3), $\mathbf{z}'_{n}$, L-ReLU \\ \hline \hline

\multirow{2}{*}{\tabincell{c}{4\\ 32$\times$32}} 
& FC, IN, CONV(512,512,3), $\mathbf{z}'_{n}$, L-ReLU 
& FC, IN, CONV(256,256,3), $\mathbf{z}'_{n}$, L-ReLU 
& FC, IN, CONV(128,128,3), $\mathbf{z}'_{n}$, L-ReLU\\ \cline{2-4} 

& FC, IN, CONV(512,512,3), $\mathbf{z}'_{n}$, L-ReLU 
& FC, IN, CONV(256,512,3), $\mathbf{z}'_{n}$, L-ReLU 
& FC, IN, CONV(128,256,3), $\mathbf{z}'_{n}$, L-ReLU \\ \hline \hline

\multirow{2}{*}{\tabincell{c}{5\\ 64$\times$64}} 
& FC, IN, CONV(512,512,3), $\mathbf{z}'_{n}$, L-ReLU 
& FC, IN, CONV(512,512,3), $\mathbf{z}'_{n}$, L-ReLU 
& FC, IN, CONV(256,256,3), $\mathbf{z}'_{n}$, L-ReLU \\ \cline{2-4} 
& FC, IN, CONV(512,512,3), $\mathbf{z}'_{n}$, L-ReLU 
& FC, IN, CONV(512,512,3), $\mathbf{z}'_{n}$, L-ReLU 
& FC, IN, CONV(256,512,3), $\mathbf{z}'_{n}$, L-ReLU \\ \hline \hline

\multirow{2}{*}{\tabincell{c}{6\\ 128$\times$128}} 
& FC, IN, CONV(512,512,3), $\mathbf{z}'_{n}$, L-ReLU
& FC, IN, CONV(512,512,3), $\mathbf{z}'_{n}$, L-ReLU 
& FC, IN, CONV(512,512,3), $\mathbf{z}'_{n}$, L-ReLU \\ \cline{2-4} 
& FC, IN, CONV(512,512,3), $\mathbf{z}'_{n}$, L-ReLU 
& FC, IN, CONV(512,512,3), $\mathbf{z}'_{n}$, L-ReLU 
& FC, IN, CONV(512,512,3), $\mathbf{z}'_{n}$, L-ReLU \\ \hline \hline

\multirow{2}{*}{\tabincell{c}{7\\ 256$\times$256}} 
& FC, IN, CONV(512,512,3), $\mathbf{z}'_{n}$, L-ReLU
& FC, IN, CONV(512,512,3), $\mathbf{z}'_{n}$, L-ReLU 
& FC, IN, CONV(512,512,3), $\mathbf{z}'_{n}$, L-ReLU \\ \cline{2-4} 
& FC(1024,512) 
& FC, IN, CONV(512,512,3), $\mathbf{z}'_{n}$, L-ReLU 
& FC, IN, CONV(512,512,3), $\mathbf{z}'_{n}$, L-ReLU \\ \hline \hline

\multirow{2}{*}{\tabincell{c}{8\\ 512$\times$512}} 
& 
& FC, IN, CONV(512,512,3), $\mathbf{z}'_{n}$, L-ReLU 
& FC, IN, CONV(512,512,3), $\mathbf{z}'_{n}$, L-ReLU \\ \cline{2-4} 
& 
& FC(1024,512) 
& FC, IN, CONV(512,512,3), $\mathbf{z}'_{n}$, L-ReLU \\ \hline \hline

\multirow{2}{*}{\tabincell{c}{9\\ 1024$\times$1024}} 
& 
& 
& FC, IN, CONV(512,512,3), $\mathbf{z}'_{n}$, L-ReLU \\ \cline{2-4} 

& 
& 
& FC(1024,512) \\ \hline
\end{tabular}
\begin{threeparttable}
\begin{tablenotes} 
\tiny 
\item[*] 
In block 1, CONV($3,C_{in},1$) is not shown in the table, which maps the RGB channels to $C_{in}$. 
\item[*] 
In CONV($C_{in},C_{out},K$), $C_{in}$ is the input channel, $C_{out}$ is output channel, and $K$ is kernel size. Stride and padding are 1 for all CONVs. 
\item[*]
FC outputs layer-wise style vectors of $\mathbf{w}'$. IN is instance normalization, L-ReLU is the activation function of Leaky ReLU. 
\item[*]
For PGGAN, $G$ inputs $\mathbf{z} \in \mathbb{R}^{512}$. $E$ outputs its imitated vector: $\mathbf{z}' \in \mathbb{R}^{512}$. We remove all FC and $\mathbf{z}'_n$, then add FC(1024*4*4, 512) in the last block.
\end{tablenotes}
\end{threeparttable} 
\end{table*}

\begin{table*}[htbp]
\tiny
\centering
\newcommand{\tabincell}[2]{\begin{tabular}{@{}#1@{}}#2\end{tabular}} 
\caption{Architectures for BigGAN encoders}
\label{tab_model_2}
\renewcommand\arraystretch{1.5} 
\setlength{\tabcolsep}{9.9pt} 
\centering
\begin{tabular}{|c|c|c|c|}
\hline
\multirow{2}{*}{\tabincell{c}{Block ID \\ Res.}} 
& \multicolumn{3}{c|}{BigGAN Encoder ($E$)} \\ \cline{2-4} 
& 128$\times$128 & 256$\times$256 & 512$\times$512 \\ \hline \hline \hline

\multirow{2}{*}{\tabincell{c}{1\\ 4$\times$4}} 
& CBN, CONV(128,128,3), L-ReLU 
& CBN, CONV(64,64,3), L-ReLU 
& CBN, CONV(32,32,3), L-ReLU \\ \cline{2-4} 
& CBN, CONV(128,256,3), L-ReLU 
& CBN, CONV(64,128,3), L-ReLU 
& CBN, CONV(32,64,3), L-ReLU \\ \hline \hline

\multirow{2}{*}{\tabincell{c}{2\\ 8$\times$8}} 
& CBN, CONV(256,256,3), L-ReLU 
& CBN, CONV(128,128,3), L-ReLU 
& CBN, CONV(64,64,3), L-ReLU \\ \cline{2-4} 
& CBN,  CONV(256,512,3), L-ReLU 
& CBN, CONV(128,256,3), L-ReLU 
& CBN, CONV(64,128,3), L-ReLU \\ \hline \hline

\multirow{2}{*}{\tabincell{c}{3\\ 16$\times$16}} 
& CBN, CONV(512,512,3), L-ReLU 
& CBN, CONV(256,256,3), L-ReLU 
& CBN, CONV(128,128,3), L-ReLU \\ \cline{2-4} 
& CBN, CONV(512,512,3), L-ReLU 
& CBN, CONV(256,512,3), L-ReLU 
& CBN, CONV(128,256,3), L-ReLU \\ \hline \hline

\multirow{2}{*}{\tabincell{c}{4\\ 32$\times$32}} 
& CBN, CONV(512,512,3), L-ReLU 
& CBN, CONV(512,512,3), L-ReLU 
& CBN, CONV(256,256,3), L-ReLU \\ \cline{2-4} 
& CBN, CONV(512,512,3), L-ReLU 
& CBN, CONV(512,512,3), L-ReLU 
& CBN, CONV(256,512,3), L-ReLU \\ \hline \hline

\multirow{2}{*}{\tabincell{c}{5\\ 64$\times$64}} 
& CBN, CONV(512,512,3), L-ReLU 
& CBN, CONV(512,512,3), L-ReLU 
& CBN, CONV(512,512,3), L-ReLU \\ \cline{2-4} 
& CBN, CONV(512,512,3), L-ReLU 
& CBN, CONV(512,512,3), L-ReLU 
& CBN, CONV(512,512,3), L-ReLU \\ \hline \hline

\multirow{2}{*}{\tabincell{c}{6\\ 128$\times$128}} 
& CBN, CONV(512,512,3), L-ReLU 
& CBN, CONV(512,512,3), L-ReLU 
& CBN, CONV(512,512,3), L-ReLU \\ \cline{2-4} 

& FC(512$\times$4$\times$4,256), FC(256,128) 
& CBN, CONV(512,512,3), L-ReLU 
& CBN, CONV(512,512,3), L-ReLU \\ \hline \hline

\multirow{2}{*}{\tabincell{c}{7\\ 256$\times$256}} 
&  
& CBN, CONV(512,512,3), L-ReLU 
& CBN, CONV(512,512,3), L-ReLU \\ \cline{2-4} 
& 
& FC(512$\times$4$\times$4,256), FC(256,128) 
& CBN, CONV(512,512,3), L-ReLU \\ \hline \hline

\multirow{2}{*}{\tabincell{c}{8\\ 512$\times$512}} 
& 
& 
& CBN, CONV(512,512,3), L-ReLU \\ \cline{2-4} 
& 
& 
& FC(512$\times$4$\times$4,256), FC(256,128) \\ \hline 
\end{tabular}
\begin{threeparttable}
\begin{tablenotes} 
\tiny 
\item[*]
CBN is the conditional batch normalization and requires the label vector as input.
\item[*]
$G$ inputs $\mathbf{z} \in \mathbb{R}^{128}$ and $\mathbf{c} \in \mathbb{R}^{256}$, and $E$ outputs the imitated vectors: $\mathbf{z}' \in \mathbb{R}^{128}$ and $\mathbf{c}' \in \mathbb{R}^{256}$.  
\end{tablenotes}
\end{threeparttable} 
\end{table*}

\begin{figure}[htbp]
\centering 
\includegraphics[width=\linewidth]{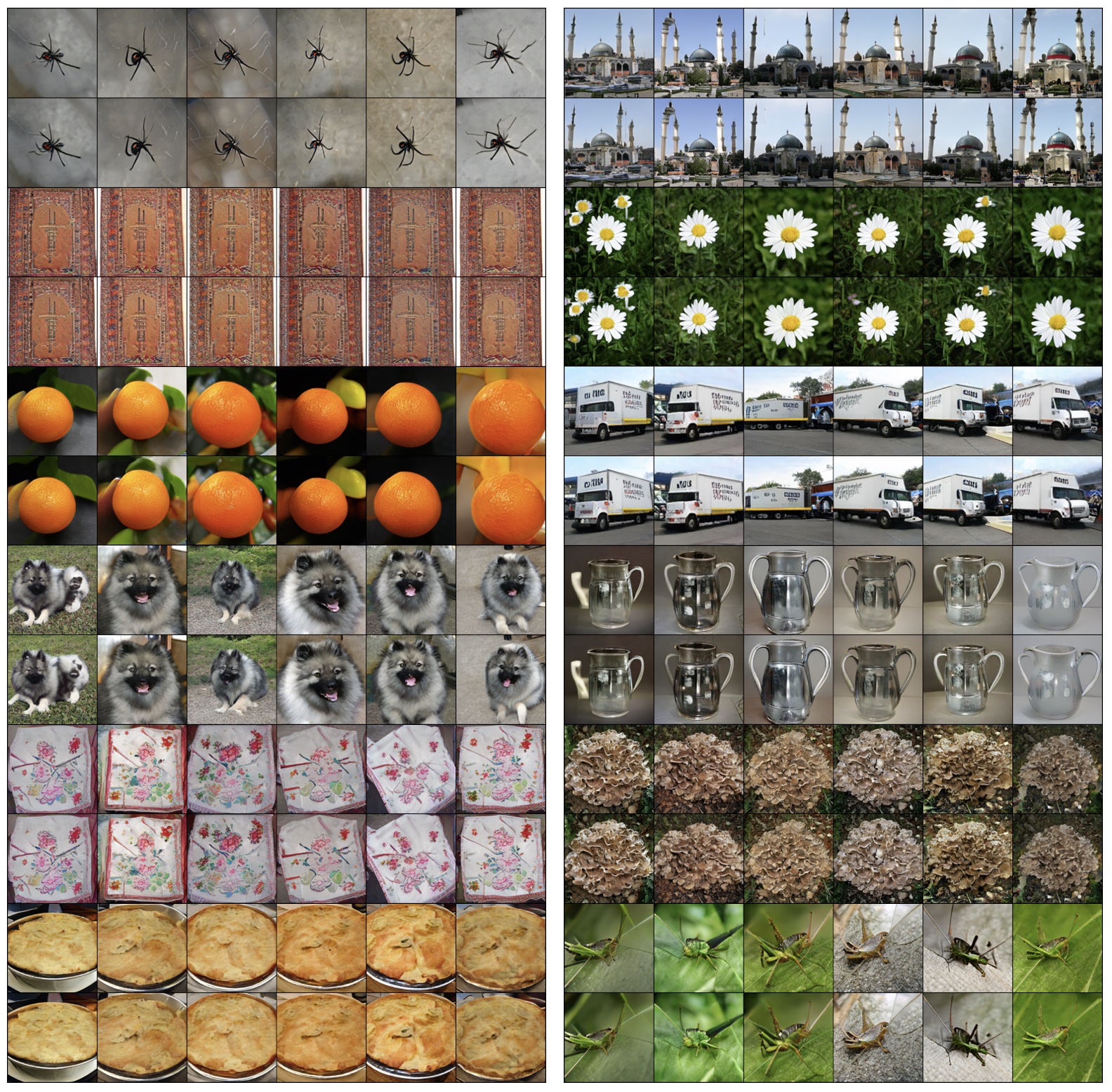} 
\caption{Visual comparison of BigGAN Inversion quality. The original BigGAN-synthesized results (upper row) and our reconstructed images (lower row) from ImageNet with a resolution of 256$\times$256. Our method successfully reproduces the BigGAN results.}
\label{fig-compare-BigGAN}
\end{figure}

\newpage
\bibliography{myReference}

\begin{thebibliography}{10}
\expandafter\ifx\csname url\endcsname\relax
  \def\url#1{\texttt{#1}}\fi
\expandafter\ifx\csname urlprefix\endcsname\relax\def\urlprefix{URL }\fi
\expandafter\ifx\csname href\endcsname\relax
  \def\href#1#2{#2} \def\path#1{#1}\fi

\bibitem{GAN}
I.~Goodfellow, J.~Pouget-Abadie, M.~Mirza, B.~Xu, D.~Warde-Farley, S.~Ozair,
  A.~Courville, Y.~Bengio, Generative adversarial nets, in: Proc. Adv. Neural
  Inf. Process. Syst. (NeurIPS), 2014, pp. 2672--2680.

\bibitem{DCGAN}
A.~Radford, L.~Metz, S.~Chintala, Unsupervised representation learning with
  deep convolutional generative adversarial networks, in: Int. Conf. Learn.
  Representations (ICLR), 2016.

\bibitem{PGAN}
T.~Karras, T.~Aila, S.~Laine, J.~Lehtinen, Progressive growing of gans for
  improved quality, stability, and variation, in: Int. Conf. Learn.
  Representations (ICLR), 2018.

\bibitem{Big-GAN}
A.~Brock, J.~Donahue, K.~Simonyan, Large scale {GAN} training for high fidelity
  natural image synthesis, in: Int. Conf. Learn. Representations (ICLR), 2019.

\bibitem{ImageNet}
J.~Deng, W.~Dong, R.~Socher, L.~Li, K.~Li, F.~Li, Imagenet: {A} large-scale
  hierarchical image database, in: Proc. IEEE Conf. Comput. Vis. Pattern
  Recognition (CVPR), 2009, pp. 248--255.

\bibitem{Style-GAN}
T.~Karras, S.~Laine, T.~Aila, A style-based generator architecture for
  generative adversarial networks, IEEE Trans. Pattern Anal. Mach. Intell.
  43~(12) (2021) 4217--4228.

\bibitem{StyleGAN2}
T.~Karras, S.~Laine, M.~Aittala, J.~Hellsten, J.~Lehtinen, T.~Aila, Analyzing
  and improving the image quality of stylegan, in: Proc. IEEE Conf. Comput.
  Vis. Pattern Recognition (CVPR), 2020, pp. 8107--8116.

\bibitem{lsun}
F.~Yu, Y.~Zhang, S.~Song, A.~Seff, J.~Xiao, {LSUN:} construction of a
  large-scale image dataset using deep learning with humans in the loop (2015).
\newblock \href {http://arxiv.org/abs/abs/1506.03365}
  {\path{arXiv:abs/1506.03365}}.

\bibitem{celeba}
Z.~Liu, P.~Luo, X.~Wang, X.~Tang, Deep learning face attributes in the wild,
  in: Proc. IEEE Int. Conf. Comput. Vis. (ICCV), 2015, pp. 3730--3738.

\bibitem{InterfaceGAN}
Y.~Shen, C.~Yang, X.~Tang, B.~Zhou, Interfacegan: Interpreting the disentangled
  face representation learned by gans, {IEEE} Trans. Pattern Anal. Mach.
  Intell. 44~(4) (2022) 2004--2018.

\bibitem{RFM}
C.~Yu, W.~Wang, H.~Li, R.~Bugiolacchi, Fast 2-step regularization on style
  optimization for real face morphing, Neural Networks 155 (2022) 28--38.

\bibitem{GAN-inversion}
W.~Xia, Y.~Zhang, Y.~Yang, J.-H. Xue, B.~Zhou, M.-H. Yang, Gan inversion: A
  survey, IEEE Trans. Pattern Anal. Mach. Intell. early access (2022) 1--17.

\bibitem{RiDDLE}
D.~Li, W.~Wang, K.~Zhao, J.~Dong, T.~Tan, Riddle: Reversible and diversified
  de-identification with latent encryptor, in: Proc. IEEE Conf. Comput. Vis.
  Pattern Recognition (CVPR), 2023.

\bibitem{InvertGAN}
A.~Creswell, A.~A. Bharath, Inverting the generator of a generative adversarial
  network, {IEEE} Trans. Neural Networks Learn. Syst. 30~(7) (2019) 1967--1974.

\bibitem{FastEncoder}
C.~Yu, W.~Wang, Fast transformation of discriminators into encoders using
  pre-trained gans, Pattern Recognition Letters 153 (2022) 92--99.

\bibitem{Perarnau2016}
G.~Perarnau, J.~van~de Weijer, B.~Raducanu, J.~M. \'Alvarez, {Invertible
  Conditional GANs for image editing}, in: NIPS Workshop on Adversarial
  Training, 2016.

\bibitem{img2stylegan}
R.~Abdal, Y.~Qin, P.~Wonka, Image2stylegan: How to embed images into the
  stylegan latent space?, in: Proc. IEEE Int. Conf. Comput. Vis. (ICCV), 2019,
  pp. 4431--4440.

\bibitem{E2Style}
T.~Wei, D.~Chen, W.~Zhou, J.~Liao, W.~Zhang, L.~Yuan, G.~Hua, N.~Yu, E2style:
  Improve the efficiency and effectiveness of stylegan inversion, IEEE
  Transactions on Image Processing 31 (2022) 3267--3280.

\bibitem{ALAE}
S.~Pidhorskyi, D.~A. Adjeroh, G.~Doretto, Adversarial latent autoencoders, in:
  Proc. IEEE Conf. Comput. Vis. Pattern Recognition (CVPR), 2020.

\bibitem{InDomainG}
J.~Zhu, Y.~Shen, D.~li~Zhao, B.~Zhou, In-domain gan inversion for real image
  editing, in: Proc. Eur. Conf. Comput. Vis. (ECCV), 2020.

\bibitem{pSp}
E.~Richardson, Y.~Alaluf, O.~Patashnik, Y.~Nitzan, Y.~Azar, S.~Shapiro,
  D.~Cohen-Or, Encoding in style: a stylegan encoder for image-to-image
  translation, in: Proc. IEEE Conf. Comput. Vis. Pattern Recognition (CVPR),
  2021.

\bibitem{e4e}
O.~Tov, Y.~Alaluf, Y.~Nitzan, O.~Patashnik, D.~Cohen{-}Or, Designing an encoder
  for stylegan image manipulation, {ACM} Trans. Graph. 40~(4) (2021)
  133:1--133:14.

\bibitem{GANspace}
E.~H{\"{a}}rk{\"{o}}nen, A.~Hertzmann, J.~Lehtinen, S.~Paris, Ganspace:
  Discovering interpretable {GAN} controls, in: Proc. Adv. Neural Inf. Process.
  Syst. (NeurIPS), 2020.

\bibitem{StyleClip}
O.~Patashnik, Z.~Wu, E.~Shechtman, D.~Cohen{-}Or, D.~Lischinski, Styleclip:
  Text-driven manipulation of stylegan imagery, in: Proc. IEEE Int. Conf.
  Comput. Vis. (ICCV), 2021, pp. 2065--2074.

\bibitem{aae}
A.~Makhzani, J.~Shlens, N.~Jaitly, I.~J. Goodfellow, Adversarial autoencoders
  (2016).
\newblock \href {http://arxiv.org/abs/abs/1511.05644}
  {\path{arXiv:abs/1511.05644}}.

\bibitem{Bi_GAN}
J.~Donahue, P.~Kr{\"{a}}henb{\"{u}}hl, T.~Darrell, Adversarial feature
  learning, in: Int. Conf. Learn. Representations (ICLR), 2017.

\bibitem{sampleGAN}
T.~{White}, Sampling generative networks, in: Proc. Adv. Neural Inf. Process.
  Syst. (NeurIPS), 2016.

\bibitem{Image-Manifold}
J.~Zhu, P.~Kr, E.~Shechtman, A.~A. Efros, Generative visual manipulation on the
  natural image manifold, in: Proc. Eur. Conf. Comput. Vis. (ECCV), 2016, pp.
  597--613.

\bibitem{BigBiGAN}
J.~Donahue, K.~Simonyan, Large scale adversarial representation learning, in:
  Proc. Adv. Neural Inf. Process. Syst. (NeurIPS), 2019, pp. 10541--10551.

\bibitem{LatentCLR}
O.~K. Y{\"{u}}ksel, E.~Simsar, E.~G. Er, P.~Yanardag, Latentclr: {A}
  contrastive learning approach for unsupervised discovery of interpretable
  directions, in: Proc. IEEE Int. Conf. Comput. Vis. (ICCV), 2021, pp.
  14243--14252.

\bibitem{perceptual}
J.~Johnson, A.~Alahi, L.~Fei-Fei, Perceptual losses for real-time style
  transfer and super-resolution, in: Proc. Eur. Conf. Comput. Vis. (ECCV),
  2016, pp. 694--711.

\bibitem{VGG}
K.~Simonyan, A.~Zisserman, Very deep convolutional networks for large-scale
  image recognition, in: Int. Conf. Learn. Representations (ICLR), 2015.

\bibitem{AlexNet}
A.~Krizhevsky, I.~Sutskever, G.~E. Hinton, Imagenet classification with deep
  convolutional neural networks, Commun. {ACM} 60~(6) (2017) 84--90.

\bibitem{SSIM}
{Zhou Wang}, A.~C. {Bovik}, H.~R. {Sheikh}, E.~P. {Simoncelli}, Image quality
  assessment: from error visibility to structural similarity, \textit{IEEE
  Trans. Image Process.} 13~(4) (2004) 600--612.

\bibitem{CAM}
B.~Zhou, A.~Khosla, {\`{A}}.~Lapedriza, A.~Oliva, A.~Torralba, Learning deep
  features for discriminative localization, in: Proc. IEEE Conf. Comput. Vis.
  Pattern Recognition (CVPR), 2016, pp. 2921--2929.

\bibitem{rethink_encoder_decoder}
Y.~Song, Rethinking the adaptive relationship between encoder layers and
  decoder layers (2024).
\newblock \href {http://arxiv.org/abs/abs/2405.08570}
  {\path{arXiv:abs/2405.08570}}.

\bibitem{ResNet}
K.~He, X.~Zhang, S.~Ren, J.~Sun, Deep residual learning for image recognition,
  in: Proc. IEEE Conf. Comput. Vis. Pattern Recognition (CVPR), 2016, pp.
  770--778.

\bibitem{IN}
D.~Ulyanov, A.~Vedaldi, V.~S. Lempitsky, Instance normalization: The missing
  ingredient for fast stylization (2016).
\newblock \href {http://arxiv.org/abs/abs/1607.08022}
  {\path{arXiv:abs/1607.08022}}.

\bibitem{aoa}
L.~Huang, W.~Wang, J.~Chen, X.~Wei, Attention on attention for image
  captioning, in: Proc. IEEE Int. Conf. Comput. Vis. (ICCV), 2019, pp.
  4633--4642.

\bibitem{CelebAMask-HQ}
C.-H. Lee, Z.~Liu, L.~Wu, P.~Luo, Maskgan: Towards diverse and interactive
  facial image manipulation, in: Proc. IEEE Conf. Comput. Vis. Pattern
  Recognition (CVPR), 2020, pp. 5549--5558.

\bibitem{understand_ssim}
J.~Nilsson, T.~Akenine{-}M{\"{o}}ller, Understanding {SSIM} (2020).
\newblock \href {http://arxiv.org/abs/abs/2006.13846}
  {\path{arXiv:abs/2006.13846}}.

\bibitem{Ptuning}
D.~Roich, R.~Mokady, A.~H. Bermano, D.~Cohen{-}Or, Pivotal tuning for
  latent-based editing of real images, {ACM} Trans. Graph. 42~(1) (2023)
  6:1--6:13.

\bibitem{adam}
D.~P. Kingma, J.~Ba, Adam: A method for stochastic optimization, in: Int. Conf.
  Learn. Representations (ICLR), 2015.

\bibitem{fid}
M.~Heusel, H.~Ramsauer, T.~Unterthiner, B.~Nessler, S.~Hochreiter, Gans trained
  by a two time-scale update rule converge to a local nash equilibrium, in:
  Proc. Adv. Neural Inf. Process. Syst. (NeurIPS), 2017, pp. 6626--6637.

\end{thebibliography}
\end{document}